\documentclass[journal,onecolumn]{IEEEtran}

\usepackage{textcomp}
\usepackage{subfigure}
\usepackage[binary-units]{siunitx}
\usepackage{algorithm,algorithmic}
\usepackage{cite}
\usepackage{booktabs}
\usepackage{bm,bbm}
\usepackage{graphicx,psfrag,float}
\usepackage{amsmath,amsthm,amsfonts}	
\usepackage{bbm}
\usepackage{color}
\usepackage{url}
\usepackage{multirow}

\begin{document}
\title{The Geodesic Distance between $\mathcal{G}_I^0$ Models and its Application to Region Discrimination}
\author{Jos\'e Naranjo-Torres,  Juliana Gambini, and Alejandro C.\ Frery~\IEEEmembership{Senior Member}

\thanks{Jos\'e Naranjo-Torres is with the Coordinaci\'on de F\'isica, Instituto Universitario de Tecnolog\'ia de Maracaibo,  Urb.\ La Floresta Av.\ 85, Edif.\ IUTM, Maracaibo, Venezuela 
 \texttt{josenaranjotorres@gmail.com}}

\thanks{Juliana Gambini is with the Depto. de Ingenier\'{\i}a Inform\'atica, Instituto Tecnol\'ogico de Buenos Aires, Av.\ Madero 399, C1106ACD Buenos Aires,  Argentina and with Depto. de Ingenier\'{\i}a en Computaci\'on, Universidad Nacional de Tres de Febrero, Pcia.\ de Buenos Aires, Argentina, \texttt{juliana.gambini@gmail.com}.}

\thanks{Alejandro C.\ Frery is with the Laborat\'orio de Computa\c c\~ao Cient\'ifica e An\'alise Num\'erica,
Universidade Federal de Alagoas, Av.\ Lourival Melo Mota, s/n, 57072-900, Macei\'o -- AL, Brazil, 
 \texttt{acfrery@gmail.com}}

}
\maketitle

\begin{abstract} The $\mathcal{G}_I^0$ distribution is able to characterize different regions in monopolarized SAR imagery. It is indexed by three parameters: the number of looks (which can be estimated in the whole image), a scale parameter and a texture parameter.
This paper presents a new proposal for feature extraction and region discrimination in SAR imagery, using the geodesic distance as a measure of dissimilarity between $\mathcal{G}_I^0$ models.
We derive geodesic distances between models that describe several practical situations, assuming the number of looks known, for same and different texture and for same and different scale.
We then apply this new tool to the problems of (i)~identifying edges between regions with different texture,  and (ii)~quantify the dissimilarity between pairs of samples in actual SAR data. 
We analyze the advantages of using the geodesic distance when compared to stochastic distances.


\end{abstract}

\section{Introduction}
\PARstart{A}{utomatic} interpretation of SAR (\textit{Synthetic Aperture Radar}) images is challenging.
It has important applications in, for instance, urban planning~\cite{6049331,5764715,6352307}, disaster management~\cite{6235981,6242371,6205771,6353568}, emergency response~\cite{7502166,6946812}, environmental monitoring and ecology~\cite{7559238,6352427,6350840,Cropdiscriminationbasedonpolarimetric}.  
One of the most important topics is the automatic discrimination of regions with different levels of texture or roughness.
  
Surface roughness is, along with the dielectric properties of the target, an important parameter.
It can be related to parameters of the imaging process as, for instance, resolution~\cite{SurfaceRoughnessScatteringHighResolutionSAR}, but it can also be described by  statistical models.

Several statistical models have been presented for 
image classification~\cite{martindenicolas2014}, 
target detection and recognition~\cite{Farrouki2005} and 
segmentation~\cite{35954,1198658,7394295,Gambiniijrs}. 
In recent years, speckled data have been modeled under the multiplicative model using the $\mathcal{G}$ family of distributions~\cite{Frery97}. 
This model is able to describe extremely textured areas better than the $\mathcal{K}$ distribution~\cite{6985522,MejailJacoboFreryBustos:IJRS}. 
Gao~\cite{Gao2010} discusses several statistical models and their relationships. 

The $\mathcal G_I^0$ distribution is able to characterize regions with different degrees of texture in monopolarized SAR imagery~\cite{Frery97}.
It depends on three parameters: 
the equivalent number of looks $L$, which is considered known or can be estimated for the whole image, 
the texture parameter $\alpha$,  
and  the scale parameter $\gamma$; the last two may vary locally.
One of the most important features of the $\mathcal{G}_I^0$ distribution is that $\alpha$ can be interpreted in terms of the roughness or texture of the observed area, which is related to the number of elementary backscatters per cell, so it is important to infer about it with quality~\cite{GambiniSC08}. 
In this work we use Maximum Likelihood estimation because its asymptotic properties are known to be excellent.

The problem of comparing two or more samples using distances or divergences,  prevails in image processing and analysis. 
Several authors employ information-theoretic measures of contrast among samples in classification~\cite{ClassificationPolSARSegmentsMinimizationWishartDistances,ClassificationFullyPolSARDiffusion-Reaction}, 
edge detection~\cite{6377288,5208318,EdgeDetectionDistancesEntropiesJSTARS,gambini2015,NonparametricEdgeDetectionSpeckledImagery,7562415}, saliency detection~\cite{6947510}  and 
despeckling~\cite{TorresPolarimetricFilterPatternRecognition}.  In the article~\cite{6410023}, the authors use the concept of Mutual Information from the Information Theory, to the glacier velocity monitoring by analyzing temporal PolSAR images. In~\cite{7326080} a  summary of the dissimilarity measurements for PolSAR image interpretation, is presented; in this paper the authors discuss and compare the Wishart distance, the Bartlett distance and the $h-\phi$ family stochastic distances. 
In~\cite{Conradsen2003} a test statistic for equality of two Hermitian positive definite matrices of a complex Wishart distribution, is presented. It can be applied as an edge detection~\cite{Schou2003}. In~\cite{ChenWang2012} a similarity test scheme is proposed for similar pixel detection. In~\cite{7010344} an active contour model using  a  ratio distance which is defined by using the probability density functions of the regions inside and outside the contours,  is proposed. In~\cite{7517404} a distance measure based on the covariance matrix of the Wishart distribution is used to crop estimation.

The Geodesic Distance~(GD) can be used to measure the difference between two parametric distributions.
It was presented by Rao~\cite{raey1945,raey1992}, and since then it has been studied by several authors~\cite{ISI:000302346300002,ISI:A1995RC74900013,AtkinsonMitchell1981}.
Berkane et al.~\cite{Berkane1997} computed a closed form for the GD between elliptical distributions,
but as such forms are not available for every pair of distributions, one has to rely on numerical solutions, e.g.  
for the GD between two Gamma models~\cite{Reverter2003155}.  

The GD has been used to solve several problems, including multivariate texture retrieval and image classification~\cite{bombrun2011,IOPORT.10013799,5946541}. 
In these articles, the authors compute a closed form for the GD between multivariate elliptical distributions under certain conditions, including the $\mathcal{G}_{\text{Pol}}^0$ polarimetric distribution~\cite{FreitasFreryCorreia:Environmetrics:03}. In~\cite{6049750} the authors present a water/land segmentation using the GD, modeling the water and the land with the Gaussian mixed model. 
Naranjo-Torres et al~\cite{Naranjo2015} presented a numerical solution for estimating the GD between samples of $\mathcal{G}_I^0$ data, under the constraint of unitary mean. 

Inspired from these works, we compute closed forms for the GD between $\mathcal{G}_I^0$ models under certain conditions and we use it as measure of contrast and in edge detection in SAR data.
These expressions were not previously available in the literature.

This new approach to measuring the separability between regions of speckled data provides a powerful tool for a number of image processing and understanding problems.

Nascimento et al.~\cite{5208318} derived a number of stochastic distances between $\mathcal G_I^0$ laws along the results by Salicru et al.~\cite{OntheApplicationsofDivergenceTypeMeasuresinTestingStatisticalHypothesessalicru} and concluded that no analytic expression is, in general, available for them.
Among these distances stems the Triangular Distance~(TD).
Gambini et al.~\cite{gambini2015} concluded that this distance outperforms others from the same class of $(h$-$\phi)$ distances (Hellinger, Bhattacharyya and R\'enyi) in a variety of situations under the $\mathcal G_I^0$ distribution.
We compare the GD and the TD, and show that the former presents a number of advantages over the latter.
Such comparison is feasible because both GD and TD can be scaled to obey the same limit distribution~\cite{OntheApplicationsofDivergenceTypeMeasuresinTestingStatisticalHypothesessalicru,
ISI:A1995RC74900013}. 

The paper unfolds as follows. 
Section~\ref{sec_SAR} recalls properties of the $\mathcal{G}_I^0$ model, including parameter estimation by maximum likelihood.
Section~\ref{geodesicdistances} presents the expressions for the GD. 
Section~\ref{TriangularDistance} shows the TD and hypothesis tests for both distances.
In Section~\ref{errorevaluation} we explain the methodology used for edge detection and to assess performance.  
Section~\ref{results} shows their application to the discrimination of simulated and actual data. 
Finally, in Section~\ref{conclu} we present conclusions and outline future work. 

\section{Speckled data and the $\mathcal G_I^0$ model}
\label{sec_SAR}

Under the multiplicative model, the return in monopolarized SAR images can be described as the product of two independent
random variables, one corresponding to the backscatter $X$ and other to the
speckle noise $Y$. In this manner, $Z=X Y$
models the return $Z$ in each pixel. 
For intensity detection, the speckle noise $Y$ is modeled as a Gamma 
distributed random variable with unitary mean whose shape parameter is the number of looks $L$.
A good choice for the distribution for the backscatter $X$ is the  reciprocal of Gamma 
law that gives rise to the $\mathcal{G}^{0}$ distribution for the return $Z$~\cite{Frery97}.
This model is an attractive choice for SAR data modeling~\cite{QuartulliDatcu:04} due to its expresiveness and
mathematical tractability.

The probability density function which characterizes the $\mathcal G_I^0(\alpha, \gamma,L)$ distribution is:
\begin{equation}
f_{\mathcal{G}_I^{0}}( z) =\frac{L^{L}\Gamma ( L-\alpha
) }{\gamma ^{\alpha }\Gamma ( -\alpha ) \Gamma (
L) }\cdot  
\frac{z^{L-1}}{( \gamma +zL) ^{L-\alpha }},%
\label{ec_dens_gI0}
\end{equation}
where $-\alpha,\gamma ,z>0$ and $L\geq 1$.
The $r$-order moments are:
\begin{equation}
E(Z^r) =\Big(\frac{\gamma}{L}\Big)^r\frac{\Gamma ( -\alpha-r )}{ \Gamma (-\alpha) }\cdot  
\frac{\Gamma (L+r )}{\Gamma (L)},
\label{moments_gI0}
\end{equation}
provided $\alpha<-r$, and infinite otherwise.

The $\alpha\in\mathbbm{R}_-$ parameter describes texture, which is related to the roughness of the target. 
Values close to zero (typically above $-3$) suggest extreme textured targets, as urban zones. 
As the value of $\alpha$ decreases, it indicates regions with moderate texture (usually $\alpha \in [-6,-3]$), as forest zones.
Textureless targets, e.g. pasture, usually produce $\alpha\in(-\infty,-6)$. 

Let $W$ be
a  $\mathcal{G}_I^0(\alpha, \gamma, L)$ distributed random variable, then 
\begin{equation}
\frac{1}{\gamma}W \sim \mathcal{G}_I^0(\alpha, 1, L),
\label{parametroescala}
\end{equation}
therefore $\gamma\in\mathbbm{R}_+$ is a scale parameter.

Given the sample  $\bm{z}=(z_1,\dots, z_n)$ of independent identically distributed random variables with common distribution $\mathcal{G}_I^0(\alpha,\gamma,L)$ with $(\alpha,\gamma) \in \Theta = \mathbbm{R}_- \times \mathbbm{R}_+ $, a maximum likelihood estimator of $(\alpha,\gamma)$ satisfies 
\begin{equation*}
(\widehat{\alpha},\widehat{\gamma}) = \arg\max_{ (\alpha,\gamma) \in \Theta}\mathcal{L}(\alpha,\gamma,L,\bm{z}),
\end{equation*}
where $\mathcal L$ is the likelihood function.
This leads to $(\widehat{\alpha},\widehat{\gamma})$ satisfying
\begin{eqnarray*}
n[\Psi^0(-\widehat{\alpha})-\Psi^0(L-\widehat{\alpha})]+\sum_{i=1}^n \ln\frac{\widehat{\gamma}+Lz_i^2}{\widehat{\gamma}}=0,\\
\frac{-n\widehat{\alpha}}{\widehat{\gamma}}-(L-\widehat{\alpha})\sum_{i=1}^n(\widehat{\gamma}+Lz_i^2)^{-1}=0,
\end{eqnarray*}
where $\Psi^0(t) = {d\ln\Gamma(t)}/{dt}$ is the digamma function. In many cases no explicit solution for this system is available and numerical methods are required. 
We used the  Broyden-Fletcher-Goldfarb-Shanno (BFGS) iterative optimization method, which is recommeded for solving unconstrained nonlinear optimization problems~\cite{Broyden65}.

\section{Geodesic Distance between $\mathcal{G}_I^0$ Models}
\label{geodesicdistances}
In this section, we present closed forms for the GD between $\mathcal{G}_I^0$ distributions. 

Let $\theta= (\theta_1,\dots,\theta_r)$, $r\geq 1$ be a parameter in $\Theta\subset\mathbbm R^r$ and 
$\{f(z\mid\theta), \theta \in \Theta\}$ a family of probability density functions of a continuous random variable $Z$. 
Under mild regularity conditions, the $(i,j)$ coordinate of the Fisher information matrix~\cite{6600931} is given by:
\begin{equation}
g_{ij}(\theta)= -E\Big(\frac{\partial^2}{\partial \theta_i \partial \theta_j} \ln f(z\mid\theta)\Big)
\label{m_Fisher}
\end{equation}
and the positive definite quadratic differential form
\begin{equation}
ds^2= \sum_{i,j=1}^d{g_{i,j}(\theta) d\theta_id\theta_j}
\label{quadratic}
\end{equation}
can be used as a measure of distance between two distributions whose parameters values are contiguous points of the parameter space.

Consider two parameters $\theta^1,\theta^2\in\Theta$, and let
$t$ be the parameter of a curve $\theta(t)\in\Theta$ which joins $\theta^1$ and $\theta^2$.
Suppose $t_1$ and $t_2$ are the values for which $\theta^\ell= \theta(t_\ell)$,  $\ell=1,2$, 
then, the GD between two probability distributions can be computed by
integrating $ds^2$ along the geodesics (locally shortest paths) between the corresponding points $\theta^1$ and $\theta^2$ and the GD $s(\theta^1,\theta^2)$ (see~\cite{Berkane1997,5946541,7551770}) is given by:
\begin{equation} 
	s(\theta^1,\theta^2)=\left|\int_{t_1}^{t_2}
	\sqrt{
	\sum_{i,j=1}^{r}{g_{i,j}(\theta)\frac{d\theta_i }{dt}\frac{d\theta_j}{dt}}
	} dt\right| .
	\label{ec_gen_DG}
		\end{equation}

The Fisher information matrix for the $\mathcal{G}_I^0(\alpha, \gamma, L)$ distribution, $g(\alpha,\gamma)$ can be computed from~\eqref{ec_dens_gI0} and~\eqref{m_Fisher}:
\begin{equation}
 \left(
\begin{array}{cc}
 \Psi ^1(-\alpha )-\Psi^1(L-\alpha ) & \frac{L}{L \gamma -\alpha  \gamma } \\
 \frac{L}{L \gamma -\alpha  \gamma } & -\frac{L \alpha }{(L-\alpha +1) \gamma ^2} \\
\end{array}
\right),
\label{mf_gi0}
\end{equation}
where $\Psi^1$ is the trigamma function. 
The quadratic differential form is obtained from~\eqref{quadratic} making $r=2$, $\theta _1=\alpha $ and $\theta _2=\gamma$:
\begin{align}
	ds^2&=g_{11}(\alpha,\gamma)d\alpha^2+(g_{12}(\alpha,\gamma)  
	+g_{21}(\alpha,\gamma))d\alpha d\gamma\nonumber \\
	&\mbox{}+g_{22}(\alpha,\gamma)d\gamma^2, \label{drao_gi0}
\end{align}
where $g_{ij}(\alpha,\gamma)$, $i,j \in \{1,2\}$ are the elements of~\eqref{mf_gi0}.

In the following we consider two special cases.
We derive the expressions for the GD between  $\mathcal{G}_I^0$ models such that 
(a)~they may differ only on the texture and the scale is known (GD between $\mathcal{G}_I^0(\alpha_1,\gamma^0)$ and $\mathcal{G}_I^0(\alpha_2,\gamma^0)$, $\gamma^0$ known), and 
(b)~the only difference between them may be the scale, and the texture is known (GD between $\mathcal{G}_I^0(\alpha^0,\gamma_1)$ and $\mathcal{G}_I^0(\alpha^0,\gamma_2)$, $\alpha^0$ known).

\subsection{Geodesic distance for known scale parameter } 
\label{sec:GDGF}

When $\gamma^0$ is known~\eqref{drao_gi0} becomes $ds^2=g_{11}(\alpha,\gamma)d\alpha^2$,
thus
\begin{equation}
s(\alpha_{1},\alpha_{2})= 
\left|\int_{\alpha_{1}}^{\alpha_{2}}\sqrt{\left[\Psi^1(-\alpha )-\Psi^1(L-\alpha )\right]}\,d\alpha\right| ,
\label{DG_gi0_c1a}
\end{equation}
and, using the following property of the trigamma function
\begin{equation*}
 \Psi^1(-\alpha )-\Psi^1(L-\alpha )=\sum _{n=1}^L (-\alpha +n-1)^{-2},
\label{g11m}
\end{equation*}
we obtain:
\begin{equation}
s(\alpha_{1},\alpha_{2})= 
\left|\int_{\alpha_{1}}^{\alpha_{2}}\sqrt{
\sum _{n=1}^L {(-\alpha +n-1)^{-2}}
}
\,d\alpha\right| .
\label{DG_gi0_c1}
\end{equation}
This equation can be solved explicitly for $L=\left\{1,2\right\}$. 

For $L=1$, the GD is given by:
\begin{equation}
	s(\alpha_{1},\alpha_{2})\Big|_{L=1}=\left|\int_{\alpha_{1}}^{\alpha_{2}}\sqrt{\frac{1}{\alpha^2}}\,d\alpha\right|=\left|\ln \frac{\alpha_2}{\alpha_1}\right|.
\label{DG_gi0_cL1}
\end{equation}
 
For $L=2$, the GD is given by:
\begin{align}
s&(\alpha_{1},\alpha_{2})\Big|_{L=2}=\left|\int_{\alpha_{1}}^{\alpha_{2}}
	\sqrt{ \frac{1}{\alpha ^2}+\frac{1}{(\alpha -1)^2}}\,
	d\alpha \right| = \nonumber\\
&\left|\ln \frac{\alpha_1^2 (\alpha_2 -1)^2  (\alpha_2 R_2 -1)((\alpha_1 -1) R_1 +1)}{\alpha_2^2(\alpha_1 -1)^2  (\alpha_1 R_1 -1) ((\alpha_2 -1) R_2 +1)} \right. \nonumber \\ 
&\left. \mbox{}+ \sqrt{2} \ln \frac{ 1+\alpha_2 (R_2 -2)- \alpha_2^2 R_2}{1+\alpha_1 (R_1 -2)- \alpha_1^2 R_1} \right|,
\label{DG_gi0_cL2}	
\end{align}
where
	\[
R_i=\sqrt{\frac{4 \alpha_i^2-4 \alpha_i+2}{(\alpha_i-1)^2 \alpha_i^2}}.
\]

\begin{figure}[hbt]
	\centering
	\subfigure[Geodesic distances, $\alpha_1=-8$, $\alpha_2 \in {[-14,-2]}$ and $L=\left\{1,2\right\}$.\label{fig:fig_dgi0_aL1a}]{\includegraphics[width=.5\linewidth]{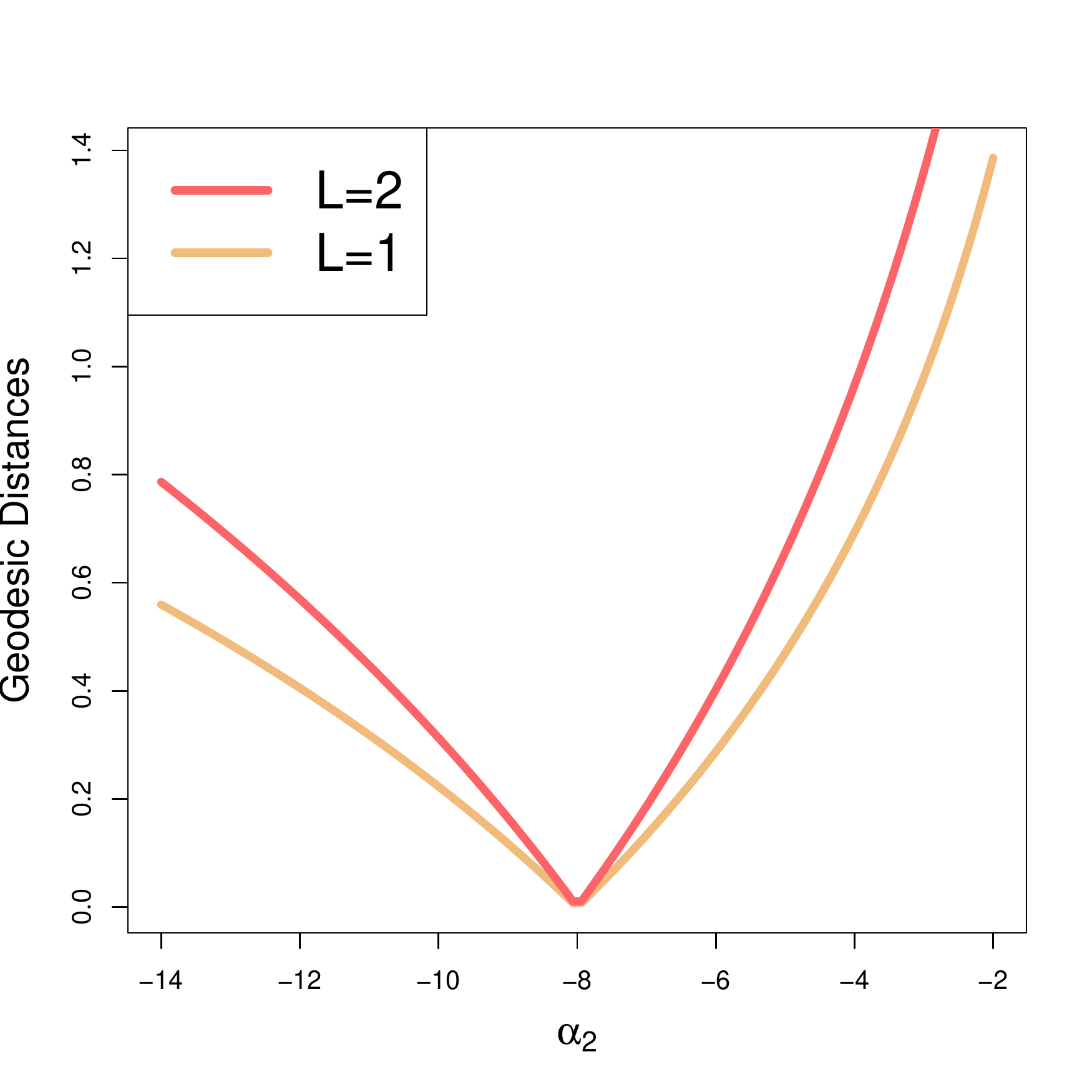}}
	\subfigure[Geodesic distances, $\alpha_1=-2$, $\alpha_2 \in {[-3.5,-1]}$ and $L=\left\{1,2\right\}$.\label{fig:fig_dgi0_aL1b}]{\includegraphics[width=.5\linewidth]{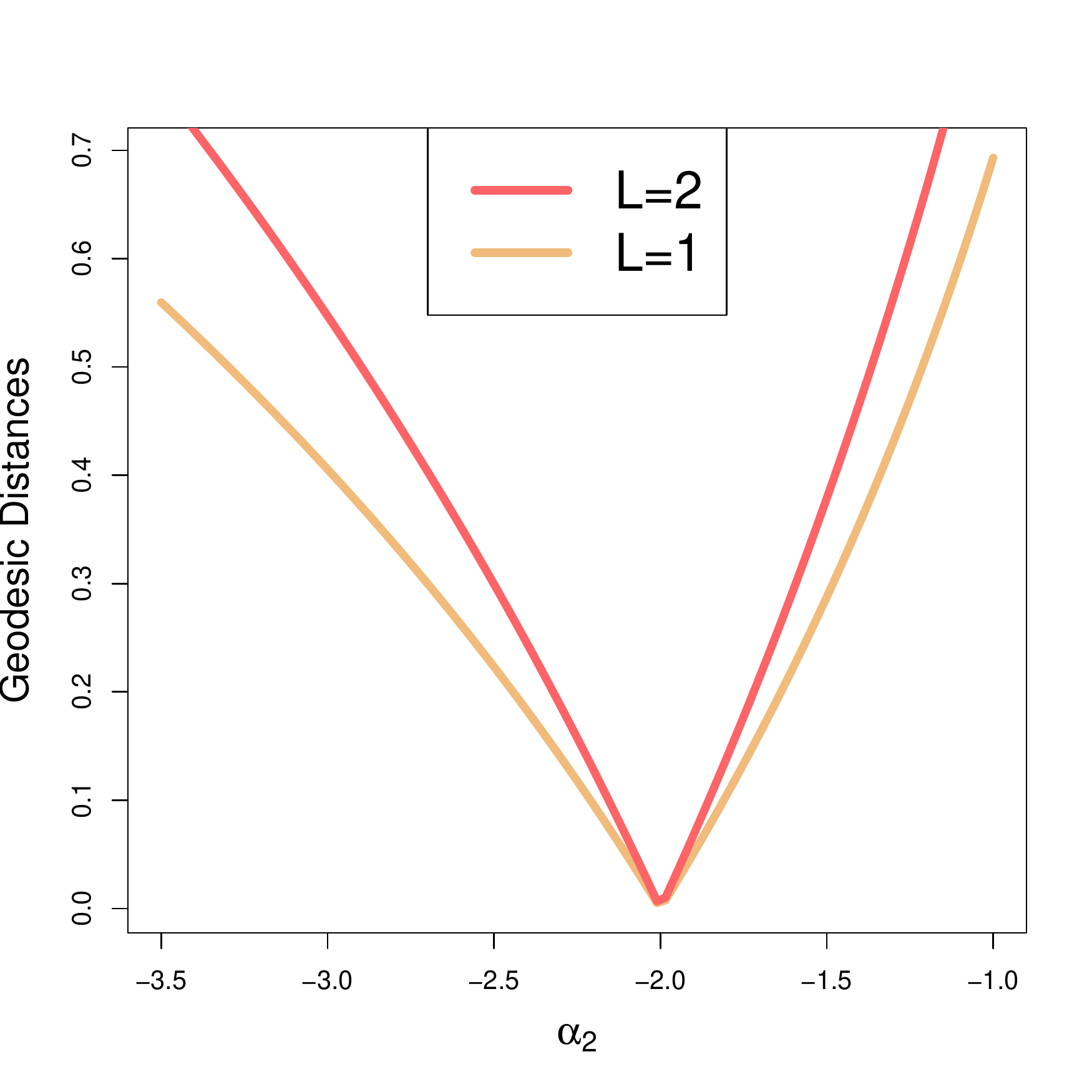}}
	\caption{Geodesic distances between  $\mathcal{G}_I^0(\alpha,1,L)$ distributions for different values of $\alpha$ and $L$.}
	\label{fig:fig_dgi0_a}
\end{figure}

Figure~\ref{fig:fig_dgi0_aL1a} shows the GD between $\mathcal{G}_I^0(\alpha,1,L)$ distributions, $L=\left\{1,2\right\}$, $\alpha_1=-8$,  $\alpha_2 \in [-14,-2]$.
Figure~\ref{fig:fig_dgi0_aL1b} shows the GDs between $\mathcal{G}_I^0(\alpha,1,L)$ distributions, $L=\left\{1,2\right\}$, $\alpha_1=-2$, $\alpha_2 \in [-3.5,-1]$. 
Notice their quasi-linear behavior, which is quite different from the quadratic-looking shapes exhibited by the stochastic distances presented in~\cite{5208318}.

Larger values of $L$
require numerical integration methods for solving~\eqref{DG_gi0_c1}.
Adaptive integration methods are recommended; cf. the Appendix.

\begin{figure}[hbt]
	\centering
	\subfigure[Geodesic distances, $\alpha_1=-8$,  $\alpha_2 \in {[-14,-2]}$ and $L=\left\{3,6,8\right\}$.\label{fig:fig_dgi0_aLNa}]{\includegraphics[width=.5\linewidth]{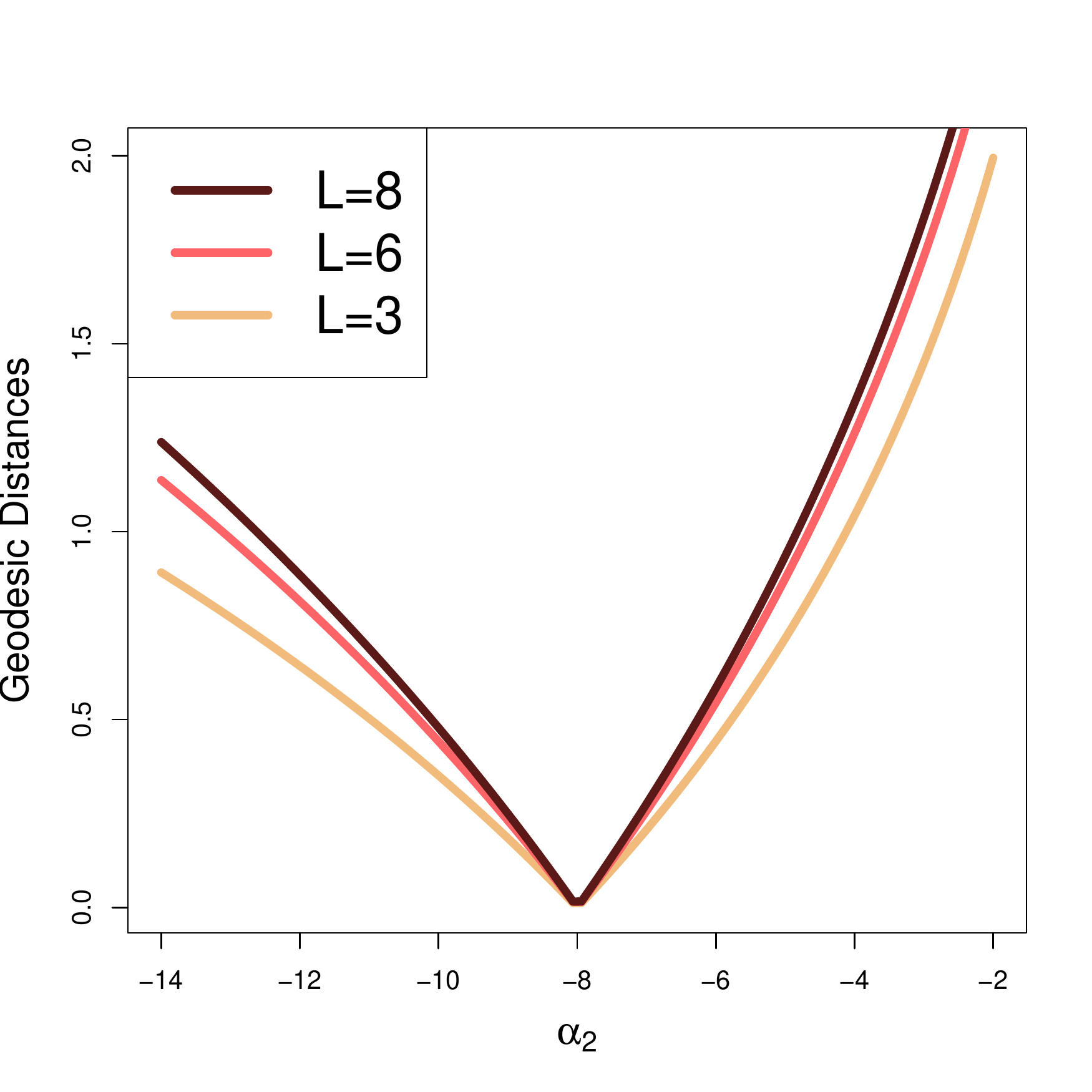}}
	\subfigure[Geodesic distances, $\alpha_1=-2$, $\alpha_2 \in {[-3.5,-1]}$ and $L=\left\{3,6,8\right\}$.\label{fig:fig_dgi0_aLNb}]{\includegraphics[width=.5\linewidth]{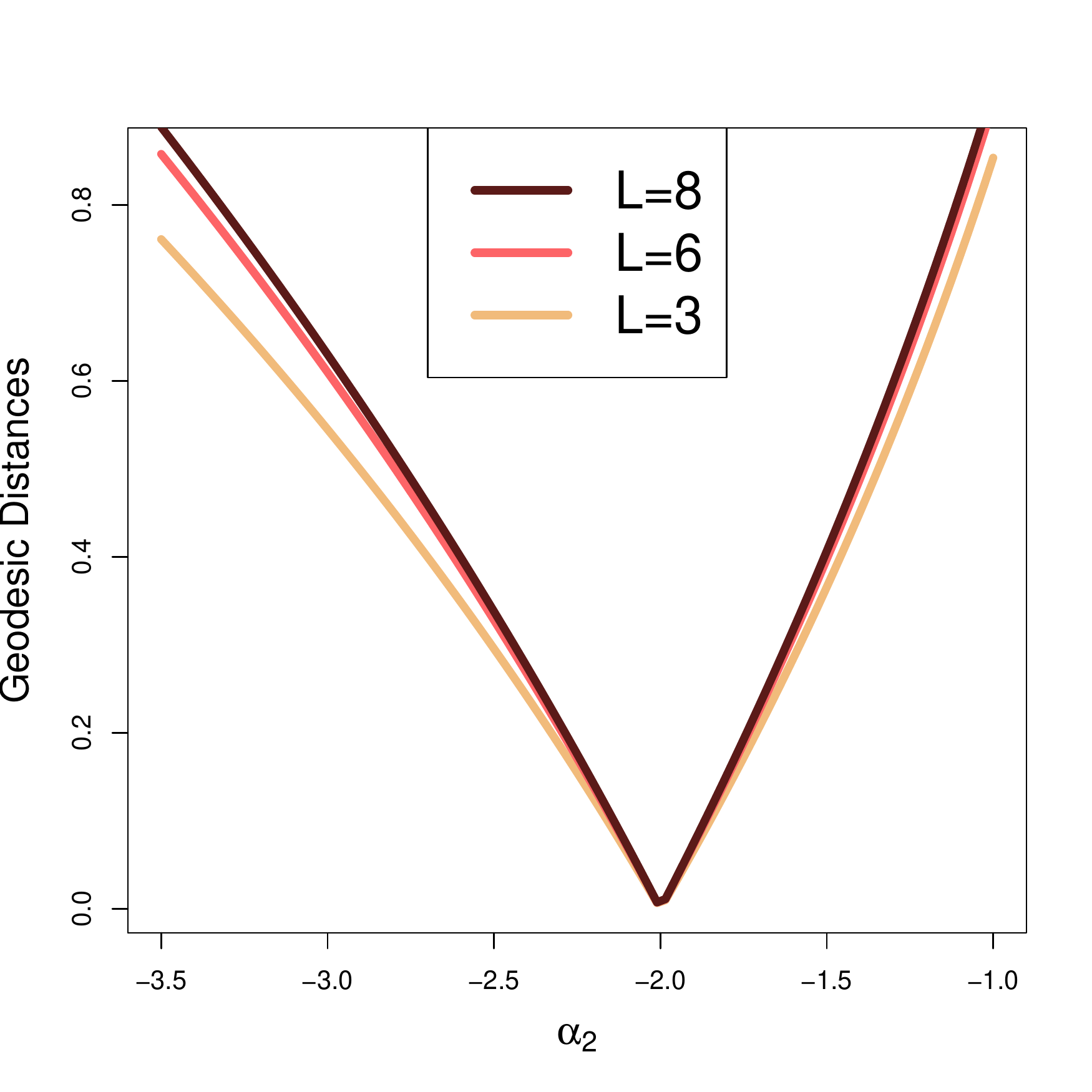}}
	\caption{Geodesic Distance computed numerically using the adaptive integration method for the $\mathcal{G}_I^0(\alpha,1,L)$ distribution.}
	\label{fig:fig_dgi0_aLN}
\end{figure}

	Figure~\ref{fig:fig_dgi0_aLNa} shows the GD between $\mathcal{G}_I^0(\alpha,1,L)$ distributions, for $L=\left\{3,6,8\right\}$, $\alpha_1=-8$,  $\alpha_2 \in [-14,-2]$.
Figure~\ref{fig:fig_dgi0_aLNb} shows the GDs between $\mathcal{G}_I^0(\alpha,1,L)$ distributions, $L=\left\{3,6,8\right\}$, $\alpha_1=-2$, $\alpha_2 \in [-3.5,-1]$.  

In all situations, it is possible to observe that, the smaller the difference between $\alpha_1$ and $\alpha_2$ is, the smaller the value of $s(\alpha_{1},\alpha_{2})$ is.
Additionally, it is observed that as the number of looks increases,  the GD $ s(\alpha_ {1}, \alpha_ {2}) $ increases, for the same values of $ \alpha_1 $ and $ \alpha_2 $.
The curve is steeper for larger values of $ \alpha $. 
This is because for small $L$, the texture parameter $\alpha $ has stronger influence on the model, 
causing higher variability than for those situations in which the signal-to-noise ratio is more favorable.

\subsection{Geodesic distance for known texture}

In this section, we compute the GD between $\mathcal{G}_I^0(\alpha, \gamma, L)$ models with $\alpha = \alpha^0$ known.
In this case~\eqref{drao_gi0} becomes $	ds^2=g_{22}(\alpha,\gamma)d\gamma^2$,
then the GD is:
\begin{align}
	s(\gamma _{1},\gamma _{2})&=\left|\int _{\gamma _{1}}^{\gamma _{2}} \sqrt {\frac{-\alpha L}{(-\alpha +L+1)\gamma ^{2}}} \, d\gamma \right| \nonumber \\ 
	&=\left|\sqrt{\frac{-\alpha L}{-\alpha +L+1}} \ln \frac{\gamma_1}{\gamma_2}\right|.
\label{DG_gi0_c2}
\end{align}
Notice that, differently from~\eqref{DG_gi0_c1}, this is a closed expression for every $L$.

\begin{figure}[hbt]
	\centering
		\subfigure[Geodesic Distance, $\gamma_1=5$, $\gamma_2\in{[1,10]}$, $\alpha=-2$ and $L=\left\{1,2\right\}$.\label{fig:dg_gi0_g2}]{\includegraphics[width=.5\linewidth]{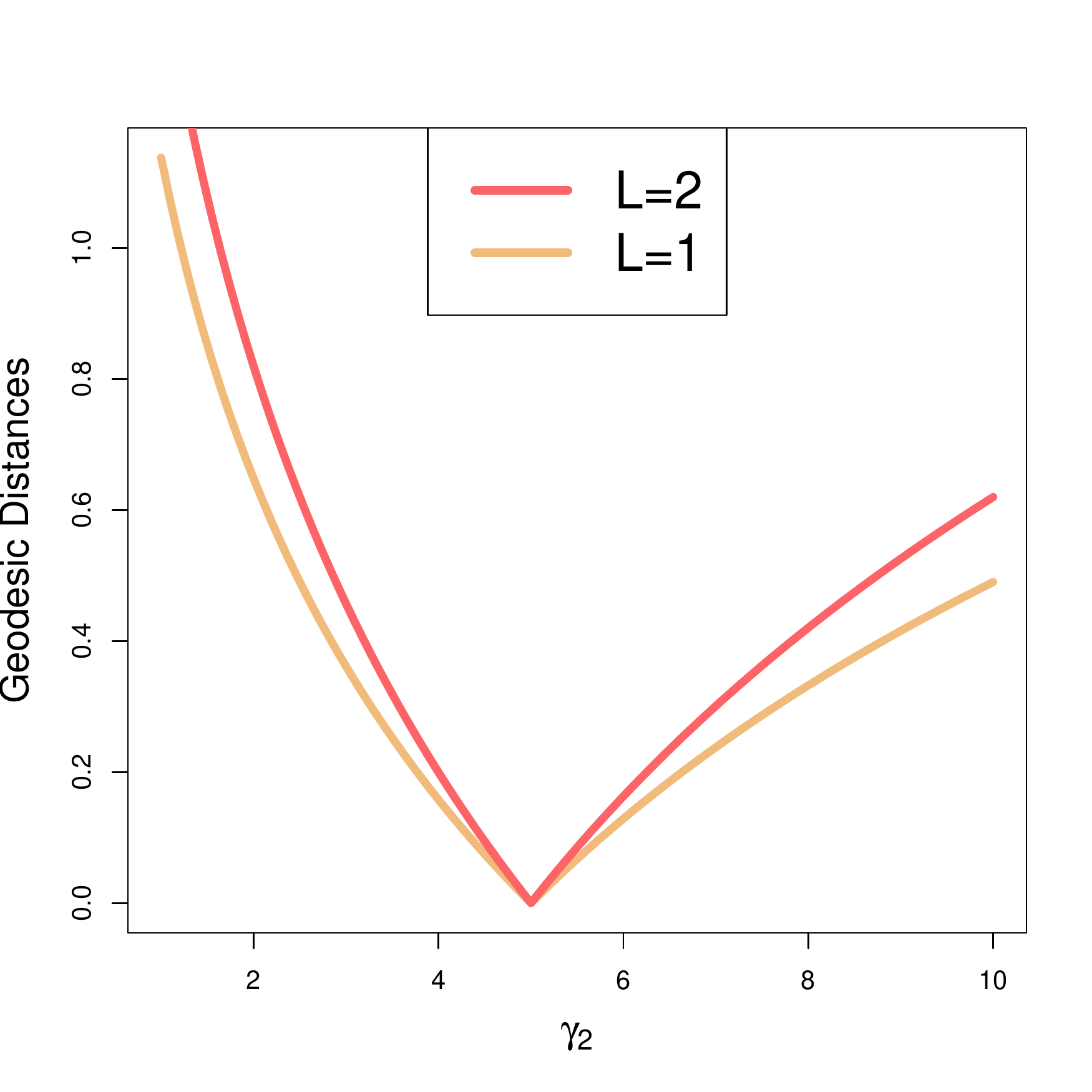}}
		\subfigure[Geodesic distance, $\gamma_1=10$, $\gamma_2\in{[1,20]}$, $\alpha=-2$ and $L=\left\{1,2\right\}$.\label{fig:dg_gi0_g1}]{\includegraphics[width=.5\linewidth]{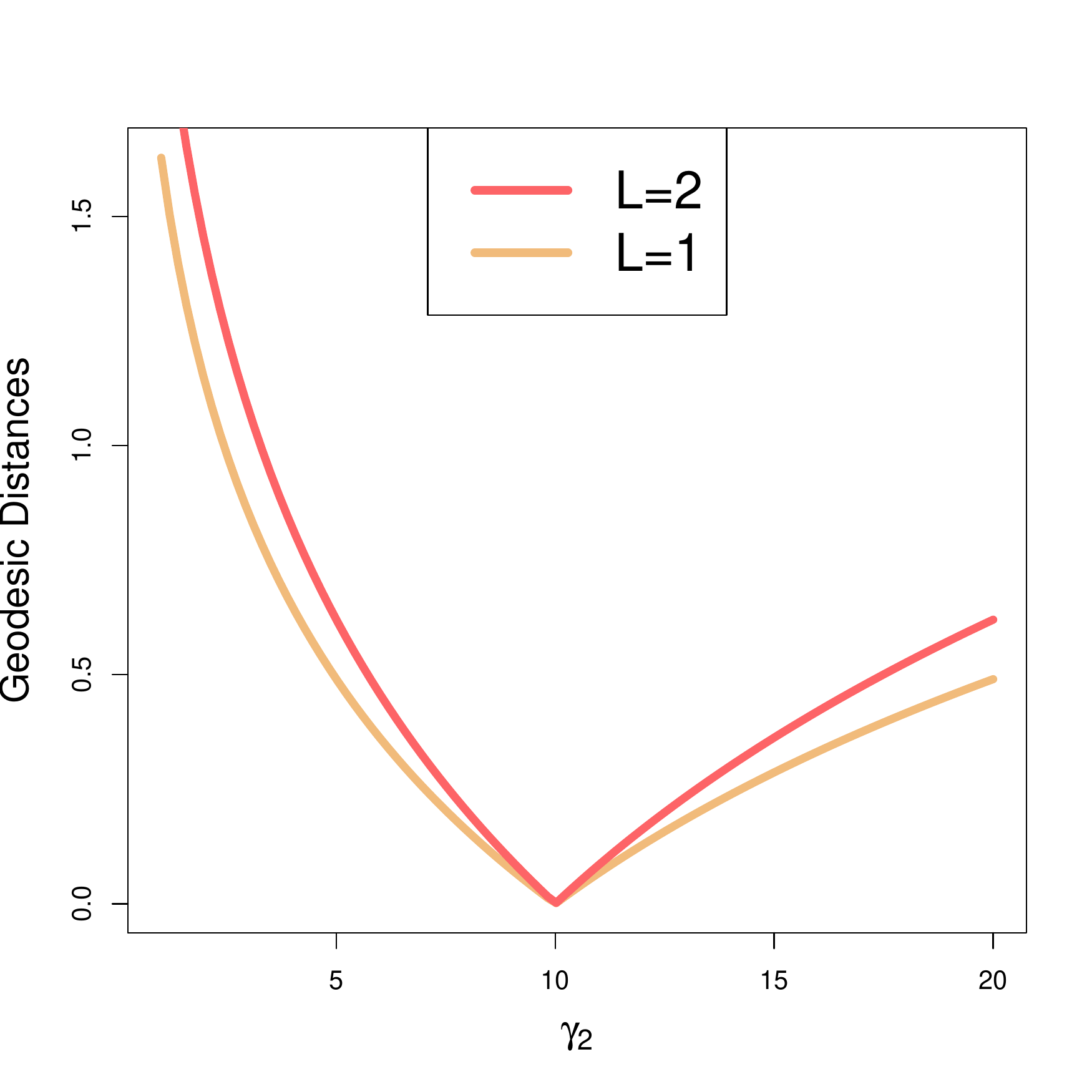}}
	\caption{Geodesic distance between  $\mathcal{G}_I^0(\alpha^0,\gamma,L)$ models.}
	\label{fig:dg_gi0_g}
\end{figure}

Figure~\ref{fig:dg_gi0_g2} shows the GDs between $\mathcal{G}_I^0(\alpha,\gamma,L)$ models with $L=\left\{1,2\right\}$, $\alpha=-2$,  $\gamma_1 = 5$, $\gamma_2 \in [1,10]$. 
Figure~\ref{fig:dg_gi0_g1} shows the GDs between models with $L=\left\{1,2\right\}$, $\alpha=-2$,  $\gamma_1 = 10$,  $\gamma_2 \in [1,20]$.  

\section{Triangular distance and Hypothesis Testing}
\label{TriangularDistance}

An important family of stochastic distances stems from the results by Salicru et al.~\cite{OntheApplicationsofDivergenceTypeMeasuresinTestingStatisticalHypothesessalicru}, 
and among them the Triangular Distance (TD) has been successfully used for SAR data analysis~\cite{gambini2015}.

Consider the densities $f_V$ and $f_W$ with common support $S$; the TD between them is given by 
\begin{equation}
d_{\text T}(f_V,f_W)=\int_{S}\frac{(f_V(z)-f_W(z))^2}{f_V(z)+f_W(z)}dz. 
\label{eq:TriangularDistance}
\end{equation}
This distance does not have explicit expression in general, and requires the use of numerical integration when applied to two $\mathcal G_I^0$ laws.


So far, we have discussed two distances between $\mathcal G^0_I$ models: GD and TD.
As such, they are not comparable and there is no possible semantic interpretation of their values.
The forthcoming results transform these distances into test statistics with the same asymptotic distribution and, with this, they become comparable and a tool for hypothesis testing.

Consider two random samples 
$\bm x = (x_1,\dots,x_m)$ and $\bm y = (y_1,\dots,y_n)$ from the 
$\mathcal{G}_I^0(\alpha_1,\gamma,L)$ and
$\mathcal{G}_I^0(\alpha_2,\gamma,L)$ laws, respectively, with $\gamma$ and $L$ known,
and the maximum likelihood estimators $\widehat\alpha_1$ and $\widehat\alpha_2$ based on them.
Compute the statistics $S_{\text{GD}}(\widehat\alpha_1,\widehat\alpha_2)$ and
$S_{\text{TD}}(\widehat\alpha_1,\widehat\alpha_2)$ given by
\begin{align}
S_{\text{GD}}(\widehat{\alpha}_1, \widehat{\alpha}_2) &= \frac{mn}{m+n}s^2(\widehat{\alpha}_1, \widehat{\alpha}_2), \label{SGD}\\
S_{\text{TD}}(\widehat{\alpha}_1, \widehat{\alpha}_2) &= \frac{2mn}{m+n}d_{\text T}(\widehat{\alpha}_1, \widehat{\alpha}_2).\label{STD}
\end{align}
Under mild regularity conditions (see~\cite{OntheApplicationsofDivergenceTypeMeasuresinTestingStatisticalHypothesessalicru,ISI:A1995RC74900013}), if the null hypothesis $H_0:\alpha_1=\alpha_2$ holds these two test statistics obey a $\chi^2_1$ distribution when $m,n\to\infty$ provided
$m(m+n)^{-1}\to\lambda\in(0,1)$.

It can be observed that the null hypothesis is equivalent to testing the hypotheses 
$s(\widehat\alpha_1,\widehat\alpha_2)=0$ and
$d_{\text T}(\widehat\alpha_1,\widehat\alpha_2)=0$.

Although these results were stated for the case of know scale and number of looks,
they are easily extended for other situations.

%

\section{Edge detection}
\label{errorevaluation}

\begin{figure}[hbt]
	\centering
		\subfigure[Densities of the $\mathcal{G}_I^0(\alpha, 1, 1)$ distribution, with different values of $\alpha$.\label{fig:FIGURAGIDensiyies}]{\includegraphics[viewport=4 16 331 302,width=.6\linewidth]{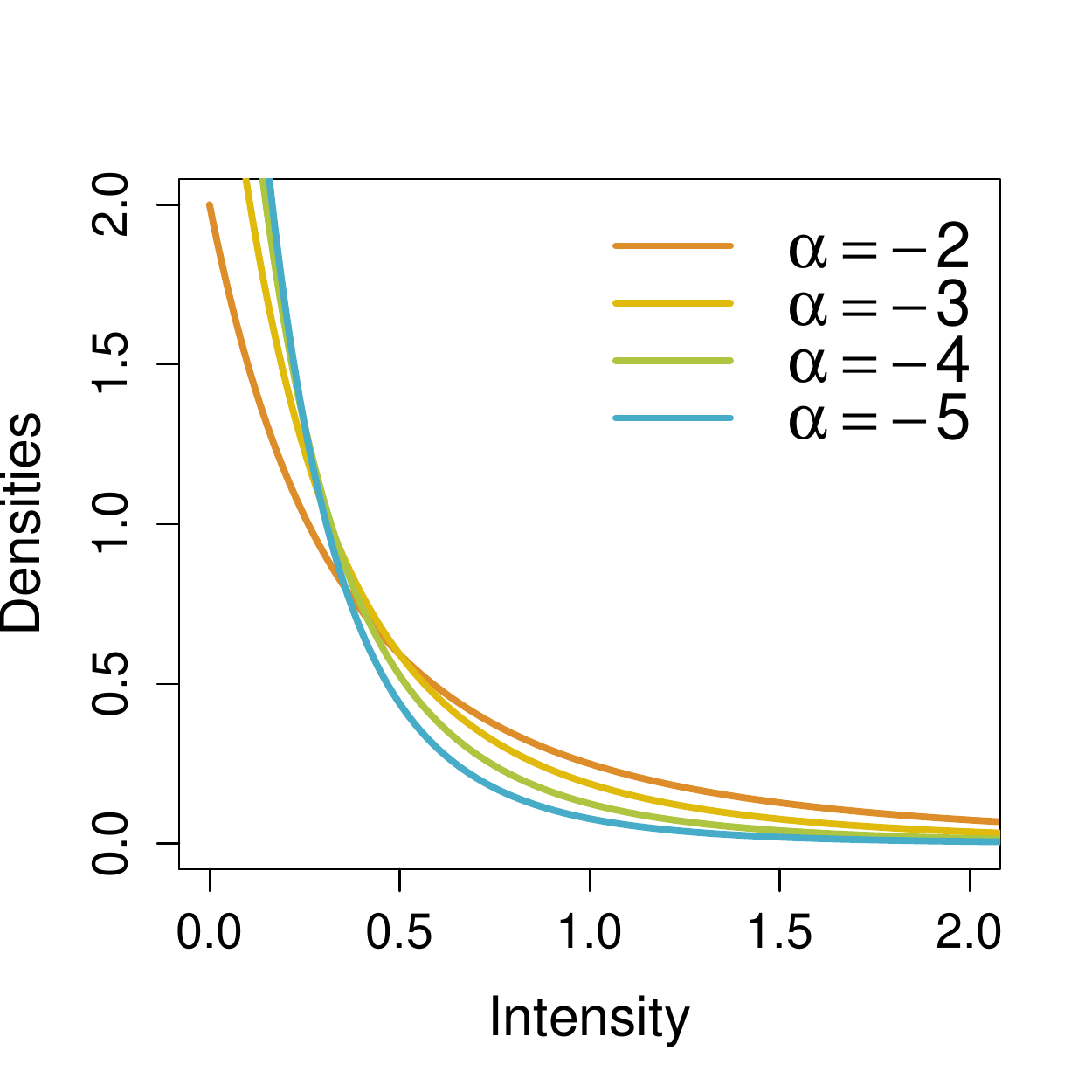}}
		\subfigure[Data from the  $\mathcal{G}_I^0(\alpha, 1, 1)$ distribution, with $\alpha\in\{-2,-3,-4,-5\}$.\label{fig:FIGURAGIDatos}]{\includegraphics[viewport=300 0 5460 1440,width=.6\linewidth]{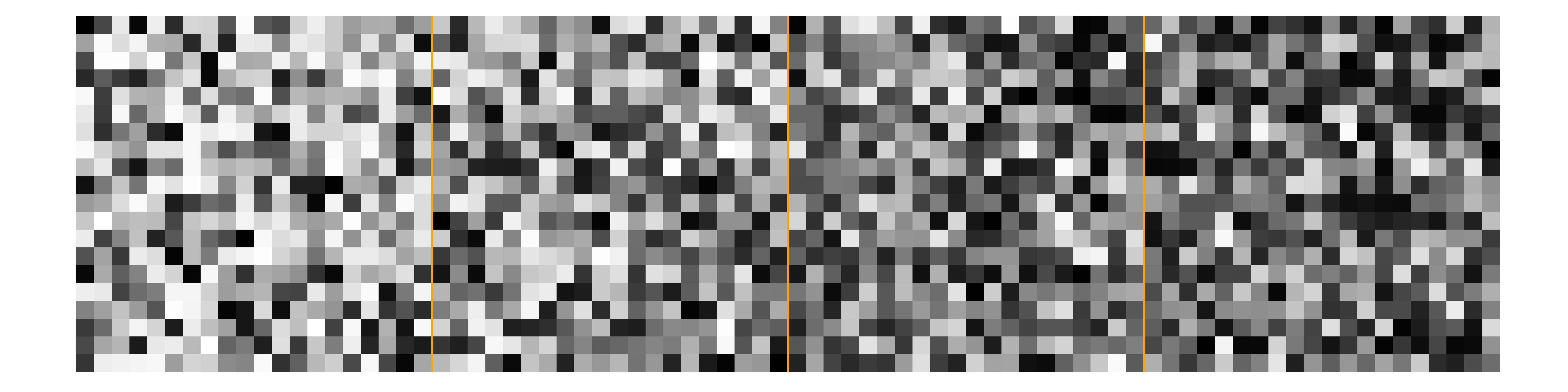}}
	\caption{The $\mathcal{G}_I^0(\alpha, 1, 1)$ model and the data it produces. }\label{fig:DifficultProblem}
\end{figure}

Discriminating targets by their roughness is a tough task when this is the only difference between them,
as presented in Fig.~\ref{fig:DifficultProblem}.
Fig.~\ref{fig:FIGURAGIDensiyies} shows the curves of the probability density of the $\mathcal G_I^0$ distribution for different values of $\alpha$ and the same scale and looks, and
Fig.~\ref{fig:FIGURAGIDatos} shows data generated with these models.  
It can be seen that both the densities and the data are hard to differentiate and, with this, the regions are difficult to distinguish.

We performed a Monte Carlo experiment to assess the discriminatory power between SAR regions through the GD and TD under the $\mathcal{G}_I^0$ distribution.

We generated $m\times n$ pixels data strips divided in halves:
one filled with samples from a $\mathcal{G}_I^0(\alpha_1, \gamma_1,L)$  
law, and the other with observations from a $\mathcal{G}_I^0(\alpha_2, \gamma_2,L)$ distribution. 

When a good edge detection algorithm is applied to this simulated image, it should give as result the position of the midpoint, which is the transition point.

\begin{figure}[hbt]
	\centering
	\includegraphics[width=.6\linewidth]{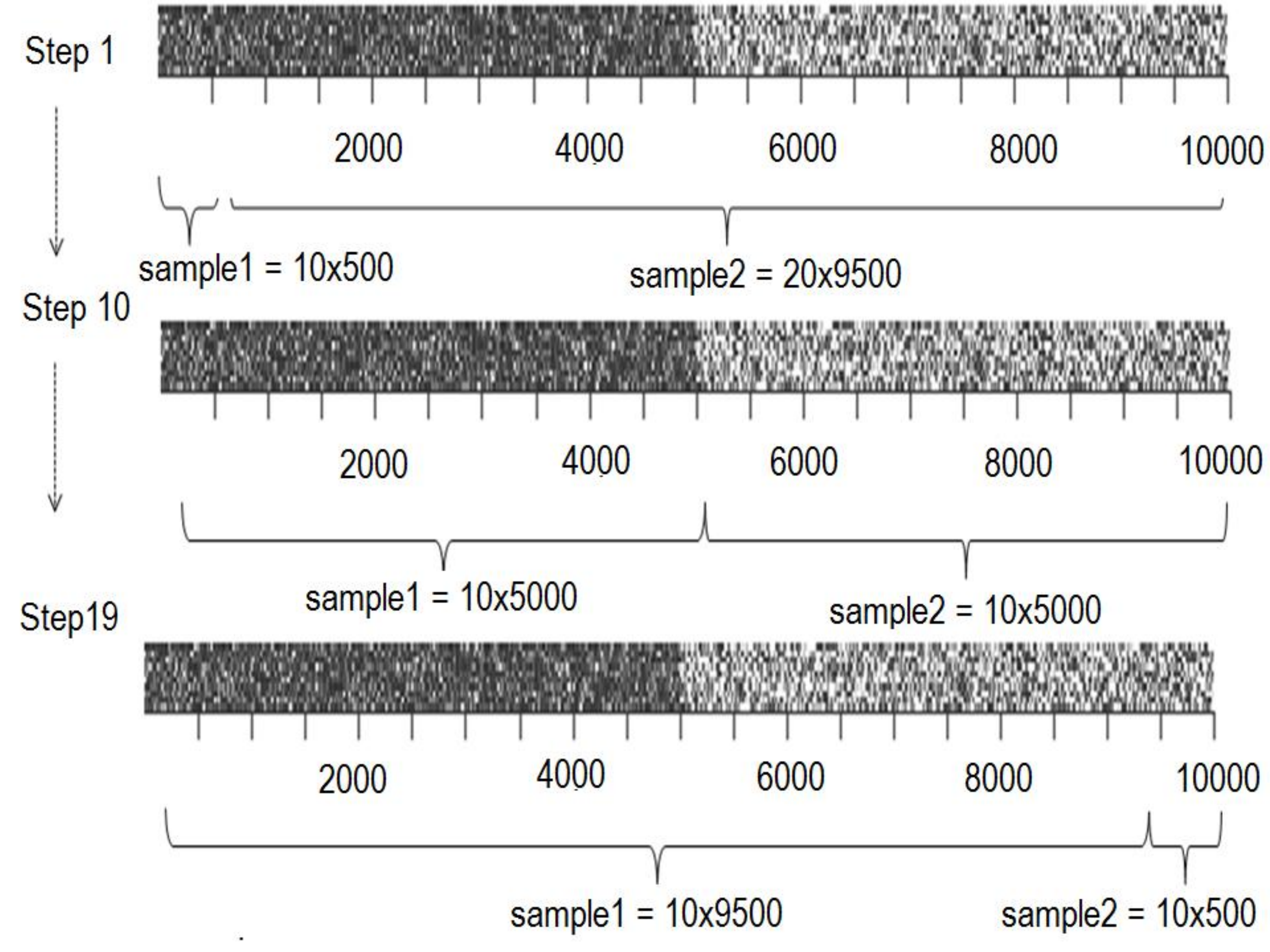}
	\caption{Simulated image and diagram illustrating the procedure to find the edge point.}
	\label{fig:Muestra}
\end{figure}

The number of elements of the sample used to estimate the parameters in the first step, $NoE$ is chosen, so $k_{top}= \left\lfloor \frac{n}{NoE}\right\rfloor-1$ estimation procedures $k=1,\dots,k_{top}$ are performed over each strip.
In each $k$ two samples are produced, $S_1(k)$ and $S_2(k)$, 
of sizes $m \times NoE*k$ and  $m \times (n- NoE*k)$ pixels, respectively. 
Figure~\ref{fig:Muestra} illustrates this idea, where $m=10$, $n= 10000$, $NoE= 500$ and $k_{top}=19$. 


Each of these two samples is used to estimate $(\alpha,\gamma)$ by Maximum Likelihood, obtaining 
$(\widehat{\alpha}_1 (k),\widehat{\alpha}_2 (k))$ and $(\widehat{\gamma}_1 (k),\widehat{\gamma}_2 (k))$. 
The data in each sample $S_i(k)$ are divided by $\widehat\gamma_i$, $i=1,2$, producing the scaled samples $S_1^*(k)$ and $S_2^*(k)$. 
According to~\eqref{parametroescala}, 
the data in $S_i^*(k)$ can be approximately described by $\mathcal{G}_I^0(\alpha_i(k), 1,L)$ distributions and, 
in doing so, 
the data on both sides have the same mean brightness.
New texture estimates are computed with these scaled data, namely $(\widehat{\alpha}_1^* (k),\widehat{\alpha}_2^* (k))$, and then the distances can be computed between the data in each sample using~\eqref{DG_gi0_c1} and~\eqref{eq:TriangularDistance}.	
These distances are then transformed into test statistics with~\eqref{SGD} and~\eqref{STD}; this transformation takes into account the different sample sizes used at every step.
Notice that, as this application uses test statistics with the same asymptotic distribution, the features and results are comparable.

We then build polygonal curves using the values of $S_{\text{GD}}$ and $S_{\text{TD}}$, as control points.
Finally, we find the positions at which these curves have the largest value. 
The method is sketched in Algorithm~\ref{Algoritmo}, where $m$ and $n$ are the number of rows and columns of the input image and the variable $NoE$ is the number of sample elements in the first step of the algorithm.  

\begin{algorithm}[hbt]
 \caption{Edge Detection through the Geodesic Distance}\label{Algoritmo}
 \begin{algorithmic}[1]
  \STATE input: $m$, $n$, $NoE$.
	\STATE Produce a strip of $m\times n$ pixels divided in halves with data from $\mathcal{G}_I^0(\alpha_1, \gamma_1,L)$  
and $\mathcal{G}_I^0(\alpha_2, \gamma_2,L)$. 
   \STATE $k_{top}= \left\lfloor \frac{n}{NoE}\right\rfloor-1$ 
	    \FOR{$k = 1,2, \dots, k_{top}$}
    		\STATE Produce samples $S_1(k)$ and $S_2(k)$ of sizes $m \times NoE*k$ and $m \times (n- NoE*k)$ pixels from the strip.
			 \STATE Estimate $(\alpha,\gamma)$ by maximum likelihood in each sample, obtaining $(\widehat{\alpha}_1 (k),\widehat{\alpha}_2 (k))$ and $(\widehat{\gamma}_1 (k),\widehat{\gamma}_2 (k))$.
			\STATE Divide the values in $S_i(k)$ by $\widehat{\gamma}_i$, obtaining $S_i^*(k)$. 
			\STATE Calculate $(\widehat{\alpha}_1^* (k),\widehat{\alpha}_2^* (k))$ with the scaled data.
			\STATE Compute $s(\widehat{\alpha}_1^*(k), \widehat{\alpha}_2^*(k))$ and $d_{\text T}(\widehat{\alpha}_1^*(k), \widehat{\alpha}_2^*(k))$ using Eq.~\eqref{DG_gi0_c1} or its particular cases and Eq.~\eqref{eq:TriangularDistance}, respectively.	 
 \ENDFOR
  \STATE Consider the array of distances between the  pairs of samples:
  $\bm s=\{s(\widehat{\alpha}_1^*(k), \widehat{\alpha}_2^*(k)), \; 1\leq k\leq k_{top}\}$  
  and
  $\bm d_{\text T}=\{d_{\text T}(\widehat{\alpha}_1^*(k), \widehat{\alpha}_2^*(k)), \;1\leq k\leq k_{top}\}$.
  \STATE Transform distances into tests statistics with~\eqref{SGD} and~\eqref{STD} obtaining the arrays $\bm S_{\text{GD}}$ and $\bm S_{\text{TD}}$ from $s$ and $d_{\text T}$, respectively.
 	\STATE Find the transition points where the test statistics are maximized:
	\begin{eqnarray*}
    \widehat{p}_s &= \arg \max_{k} \bm S_{\text{GD}},\\
    \widehat{p}_{\text T} &= \arg \max_{k} \bm S_{\text{TD}}.
	\end{eqnarray*}
 \end{algorithmic}
 \end{algorithm}
 
Notice that this procedure provides a coarse estimation of the underlying distance between the patches.
In practical applications, as those discussed by~\cite{GambiniSC08,EdgeDetectionDistancesEntropiesJSTARS,NonparametricEdgeDetectionSpeckledImagery},
a finer search strategy must be employed.

\section{Results}\label{results}

In this section, we present the results of applying the features and methods described in previous sections. 
We use simulated and actual data, and we compare the performance of the GD with respect to the TD. 
 
\subsection{Results with simulated data}\label{sec:EdgeSimulatedData}

\begin{figure}[hbt]
	\centering
		\subfigure[Mean polygonal curves for $L=1$.]{\includegraphics[width=.6\linewidth]{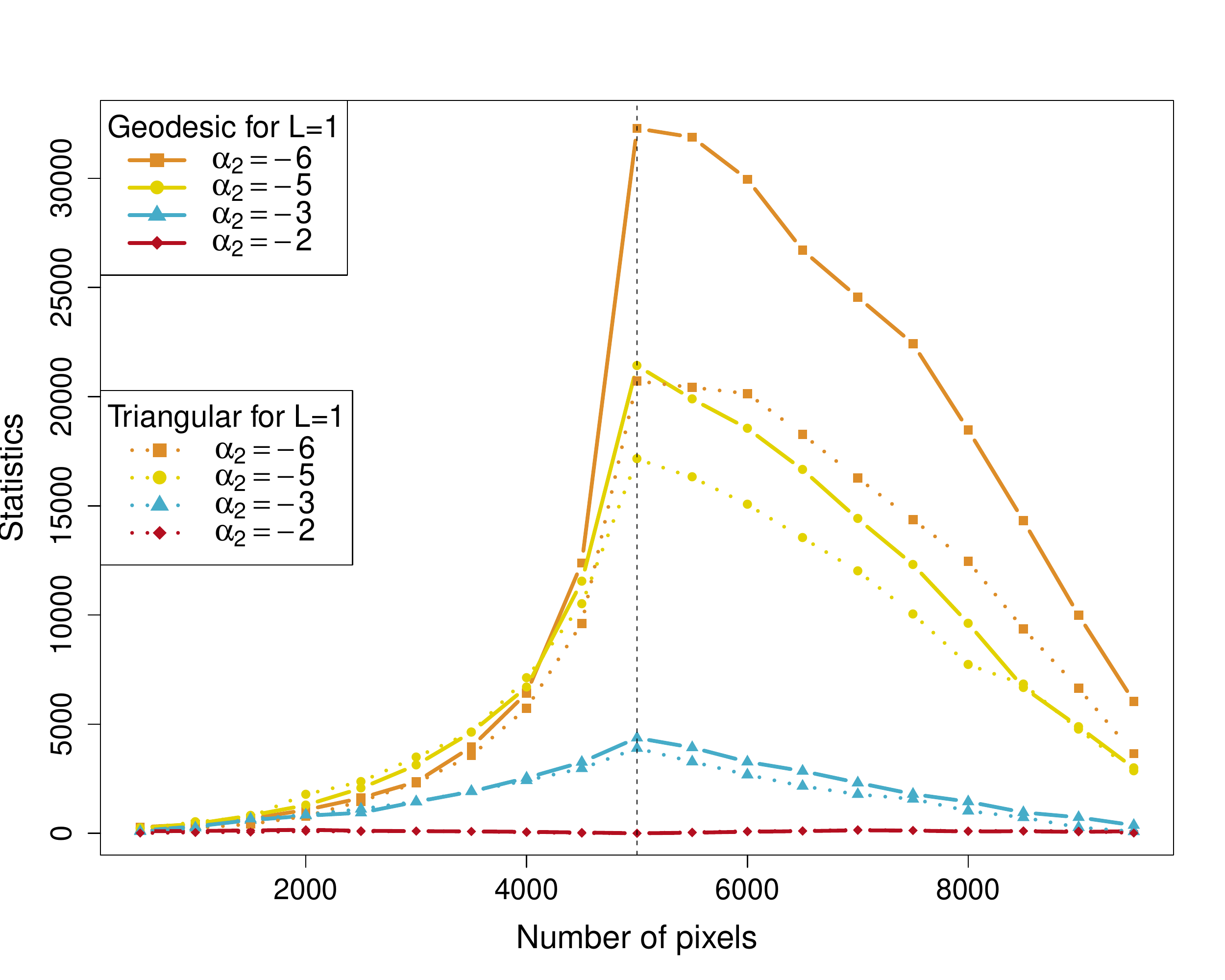}\label{fig:FIGURA2DG1}}
		\subfigure[Mean polygonal curves for $L=2$.
	]{\includegraphics[width=.6\linewidth]{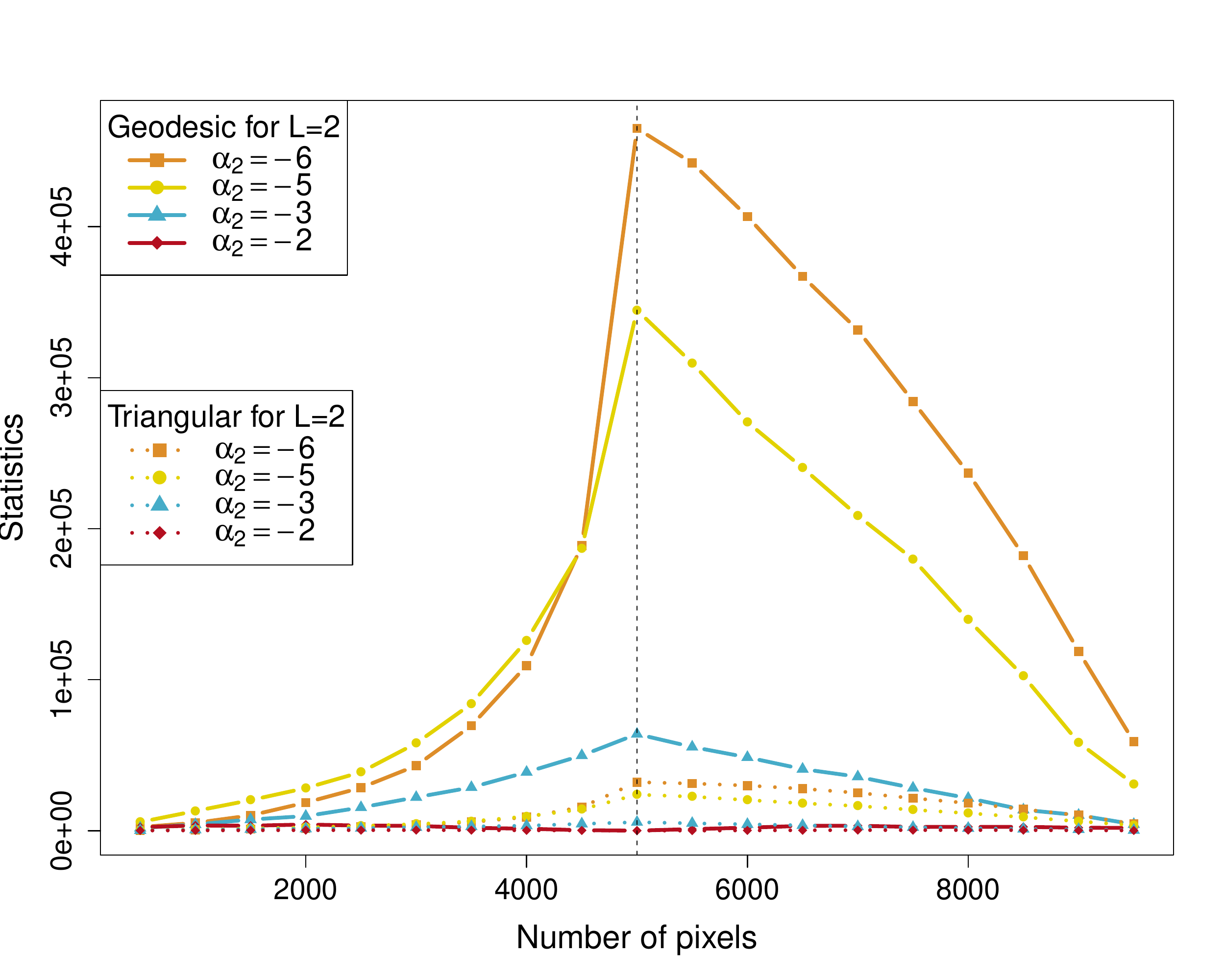}\label{fig:FIGURA2DG2}}
	\caption{Test statistics based on the Geodesic and Triangular distances between potentially mixed samples from $\mathcal{G}_I^0(-2,1,L)$ and $\mathcal{G}_I^0(\alpha_2,1,L)$ laws, with $\alpha_2=\{-2,-3,-5,-6\}$.} 
	\label{fig:FIGURA2DG}
\end{figure}

The procedure sketched in Algorithm~\ref{Algoritmo} was repeated $1000$ times over strips obtained with different texture and looks parameters, independently.
Fig.~\ref{fig:FIGURA2DG} shows the mean polygonal curves obtained in each situation by computing $S_{\text{GD}}$ and $S_{\text{TD}}$.


It can be seen that the transition, which is indicated with a vertical dotted line, can be estimated from these polygonal curves as it is consistently the largest value.

The evidence provided by the $S_{\text{GD}}$ is more conclusive that coming from  $S_{\text{TD}}$, as can be seen comparing the solid and dotted lines of same color.
In particular, the TD provides little or no evidence to find the edge between $\alpha_1=-2$ and $\alpha_2=-3$ and $L=2$.

In addition, the larger the difference between $\alpha_1$ and $\alpha_2$ is, 
the larger are the values of the maxima.

The procedure is, thus, able to identify the position of the transition point between areas with different texture and same mean brightness, a difficult task in edge detection.

We also studied the case where there is no edge, i.e., $\alpha_1=\alpha_2=-2$.
The polygonal curves are shown in dark red, and they are, for both distances and looks, the smallest.
These curves do not exhibit either global or local maxima that may lead to erroneous identification of edges.
The proposal is, thus, overall not prone to producing false alarms. 



%
%
%

\subsection{Empirical $p$-values}

Although the asymptotic distribution of~\eqref{STD} and~\eqref{SGD} is known, 
it is also valuable to know about their finite size sample behavior.

\begin{figure}[hbt]
	\centering
	\includegraphics[width=.5\linewidth]{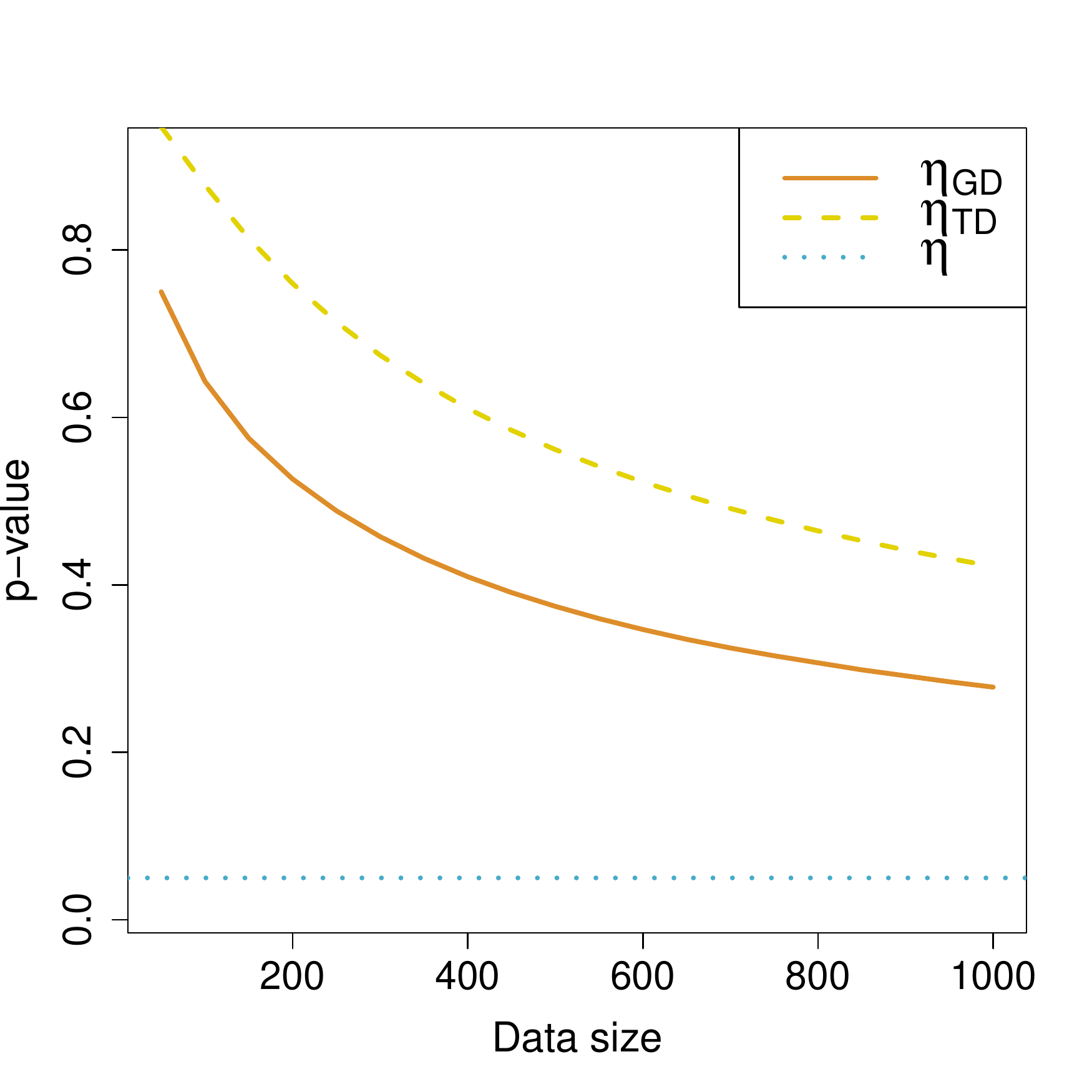}
	\caption{Empirical $p$-values for different sample sizes.}
	\label{fig:p-valor}
\end{figure}

Two samples of the same distribution and same size were generated, and the number of cases for which
the test statistics produced values larger than the critical one at \SI{5}{\percent},
namely, $3.841459$.
This was repeated \num{5000} times, and the results are shown in Fig.~\ref{fig:p-valor}.

Both tests tend to reject less than expected, since the empirical $p$-values are larger than the asymptotic one.
The test based on the GD, $S_{\text{GD}}$ converges faster to $\eta = 0.05$ than the one that relies on the TD.
It is also consistently closer to the asymptotic value.

\subsection{Results with actual data}\label{sec:ActualData}

In the following, we show the results of applying the proposed technique to an E-SAR  image~\cite{Horn1998} from the surroundings of Munich, Germany.
We employed L band, HH polarization in single-look complex format to produce intensity data.
The original resolution is \SI{2.2 x 3.0}{\squared\meter} (range $\times$ azimuth).
The intensity of each pixel is henceforth referred to as ``full resolution'' data;
the subsampled average of four contiguous pixels is ``reduced resolution''.
Theoretically, the former has one look while the latter has four looks.

\begin{figure}[hbt]
	\centering
	\subfigure[SAR image and the region used in Figs.~\ref{fig:BorderDetection} and~\ref{fig:BorderDetectionTouzi}. \label{fig:OriginalImage}]{\includegraphics[width=.3\linewidth]{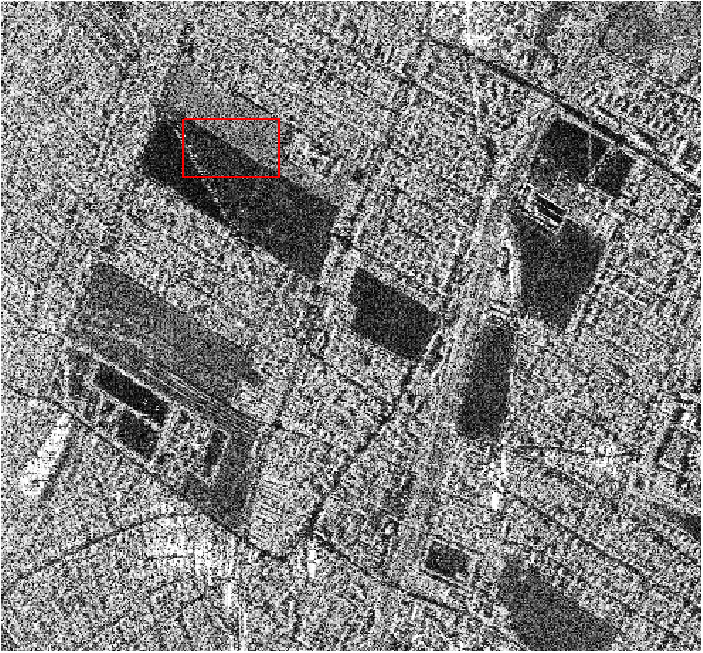}}
	\subfigure[The corresponding optical image from Google Maps. \label{fig:OriginalMap}]{\includegraphics[width=.3\linewidth]{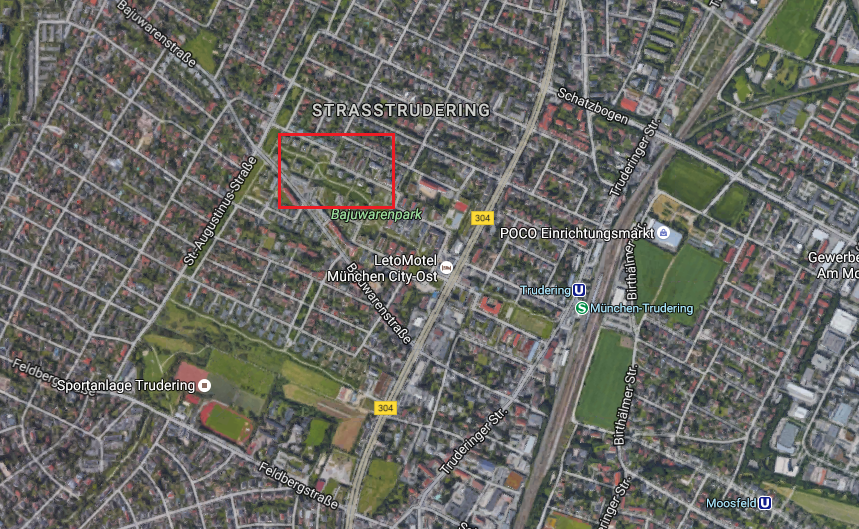}}\\
\subfigure[Edge points found over each line of the image with the GD and TD. \label{fig:BorderDetection}]{\includegraphics[width=.6\linewidth]{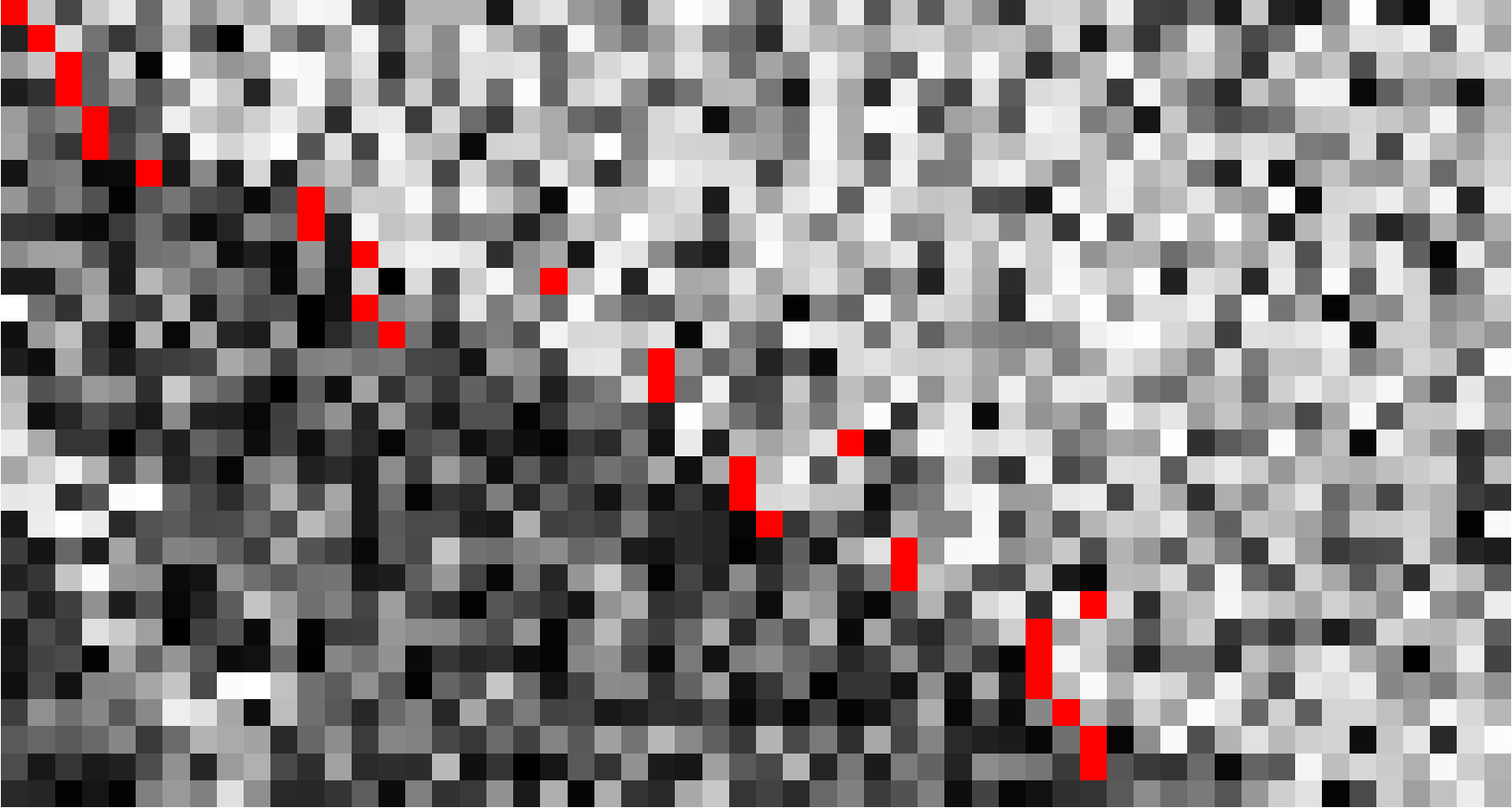}}
\subfigure[Edge points found in the image with lower resolution, with the GD. \label{fig:BorderDetectionRed}]{\includegraphics[width=.6\linewidth]{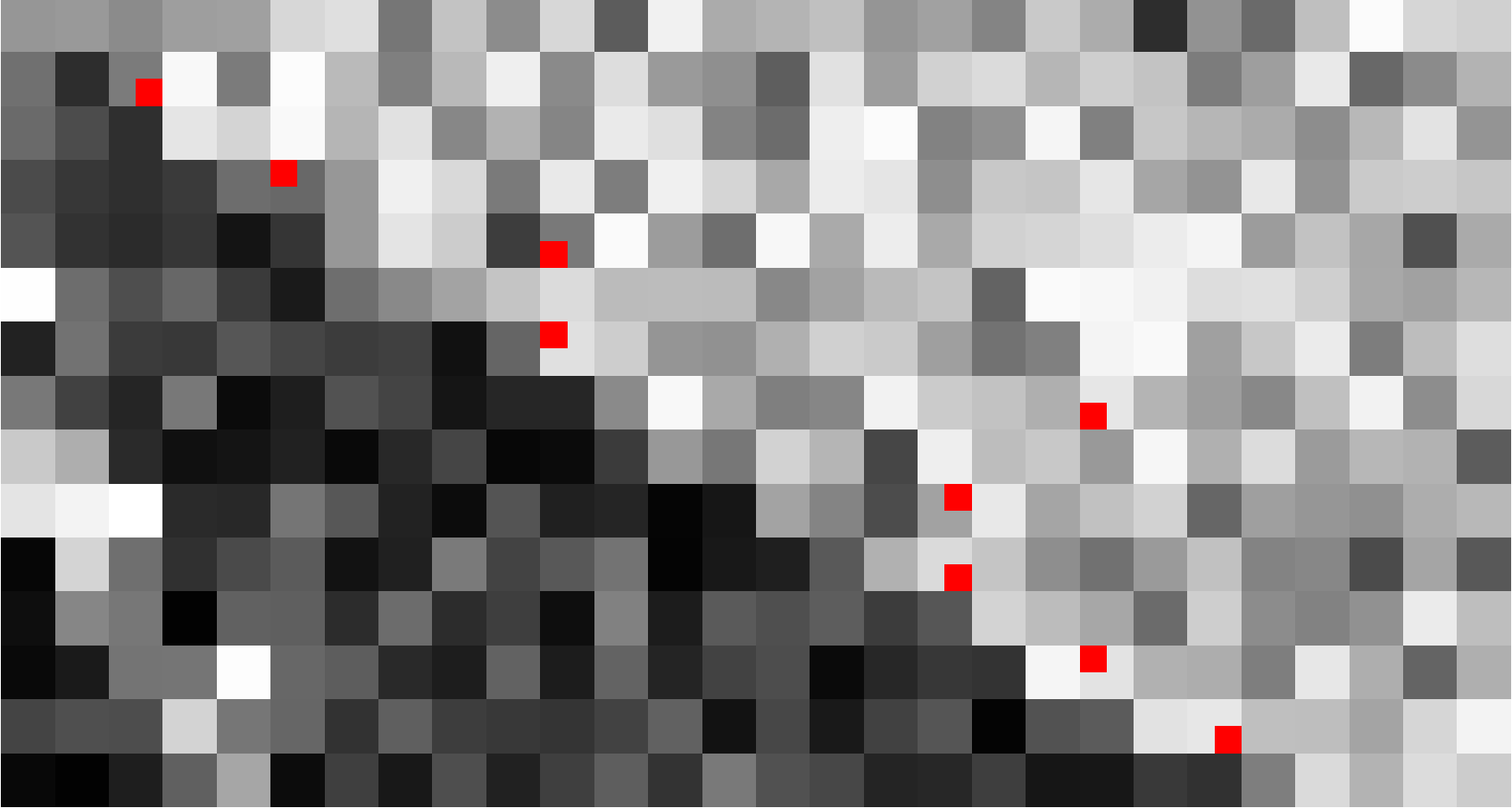}}

\caption{Results of applying the edge detectors to actual data. }
\label{fig:EdgeDetectors}
\end{figure}

Figure~\ref{fig:OriginalImage} shows the SAR image, and presents the area where the edge detection was performed. 
Figure~\ref{fig:OriginalMap} is the corresponding optical image from Google Maps. 
Figure~\ref{fig:BorderDetection} shows the result of applying the edge detector with both GD and TD to
equally-spaced horizontal strips of side $3$.
It is remarkable that both led to the same edge points, differing only in the time it took to calculate them,  except when the integral of the TD does not converge and therefore it returns no result. 
The computation time of TD is $70$ times larger the computation time of GD, and it fails to converge with small samples. 
Figure~\ref{fig:BorderDetectionRed} shows the edge detection using only GD, to the reduced resolution image. 
It can be seen that two points are shifted from the right place, but the result is acceptable.
 
\begin{figure}[hbt]
	\centering
\subfigure[Result of the Touzi edge detector, full resolution data.\label{fig:BorderDetectionTouzi}]{\includegraphics[width=.6\linewidth]{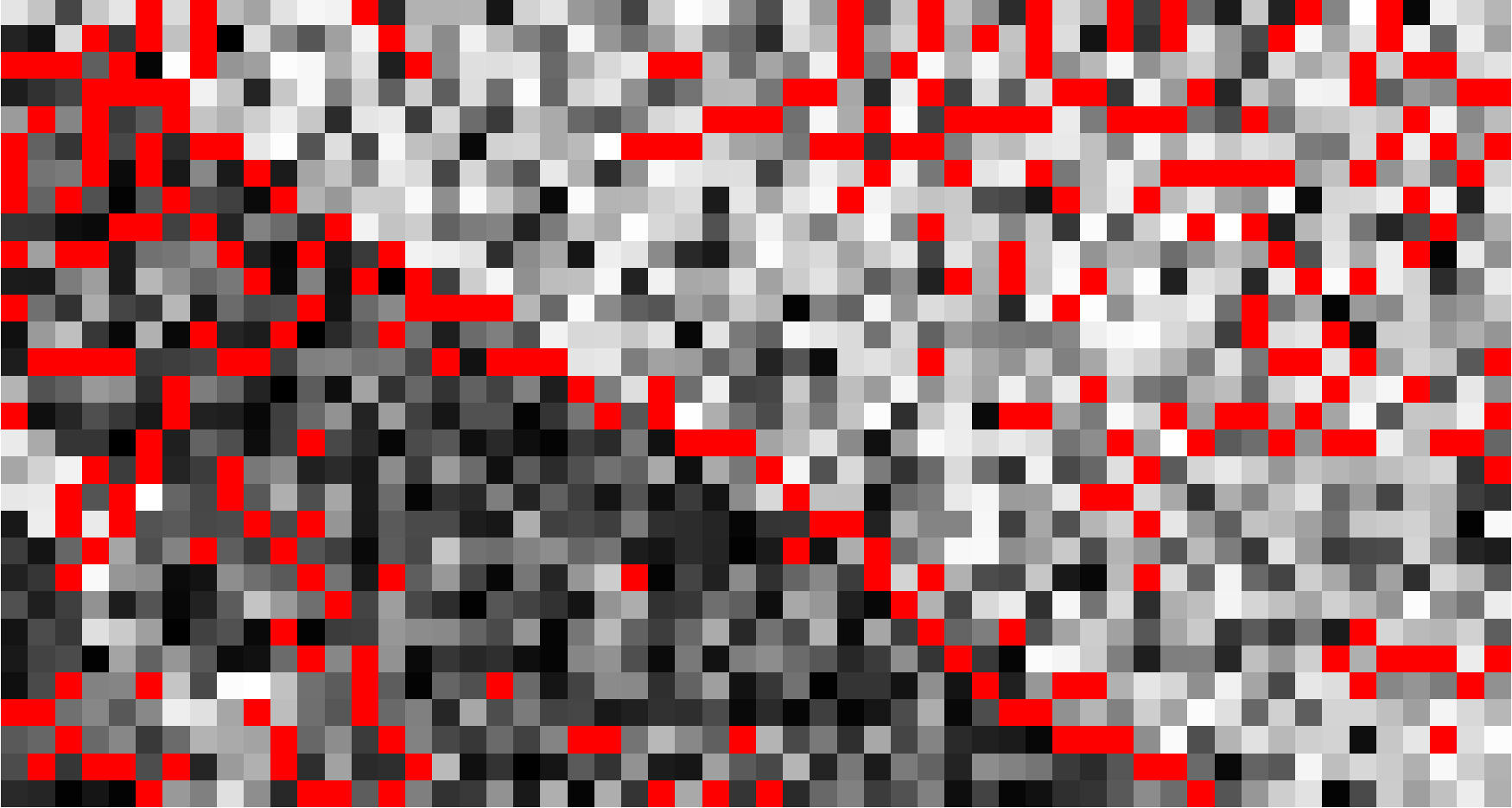}}
\subfigure[Result of the Touzi edge detector, reduced resolution data.\label{fig:BorderDetectionTouziRed}]{\includegraphics[width=.6\linewidth]{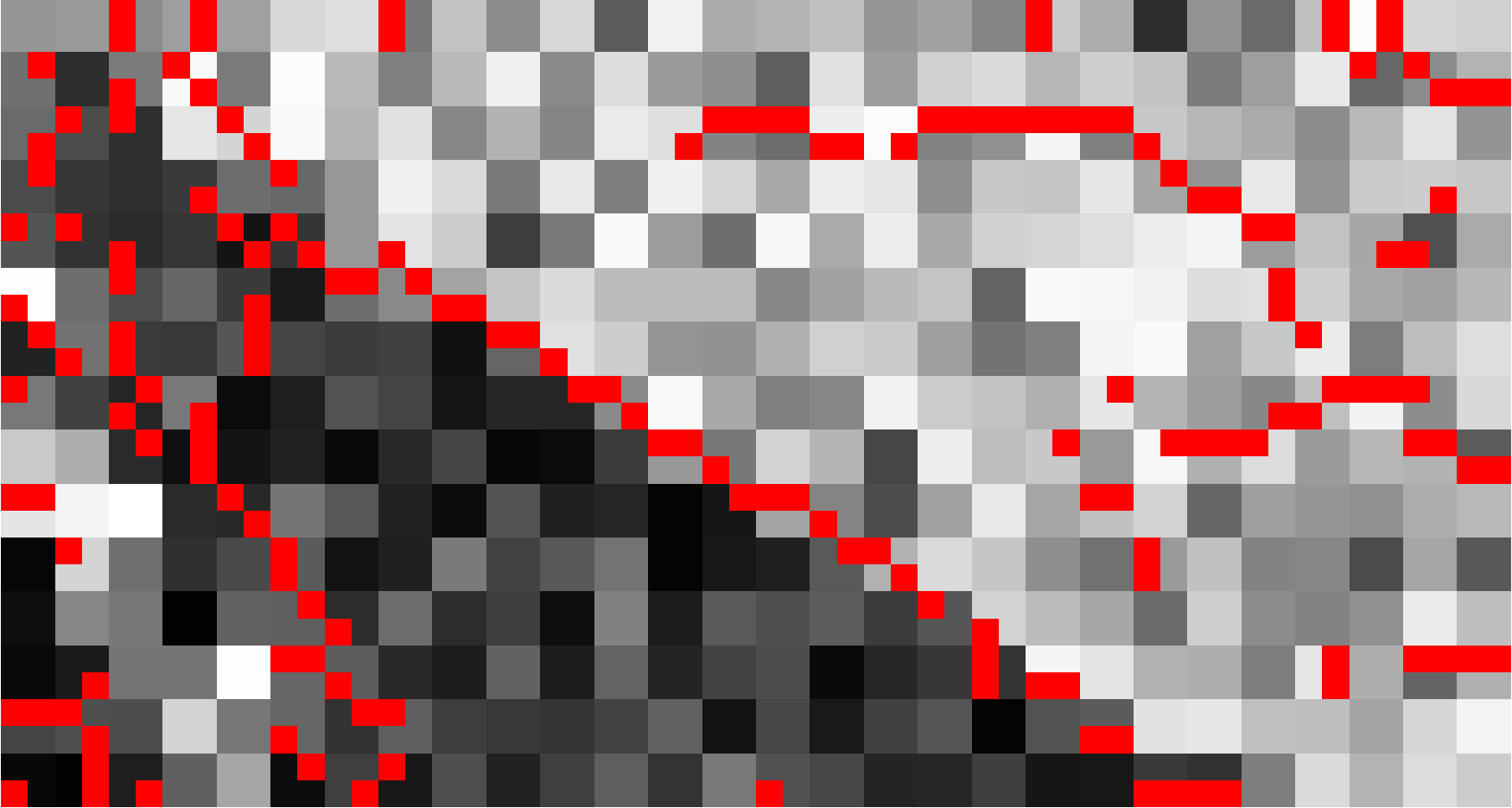}}
	\caption{Results of applying the Touzi edge detectors to actual data. }
	\label{fig:EdgeDetectorsTouzi}
\end{figure}

In order to show the advantage of our method, the ROA edge detector proposed by Touzi et al.~\cite{touzi88} was applied to the same data of Figs.~\ref{fig:BorderDetection} and~\ref{fig:BorderDetectionRed}. 
The results are shown in Figs~\ref{fig:BorderDetectionTouzi} and~\ref{fig:BorderDetectionTouziRed}. 
The parameters used are: 
(i)~increasing window size: $3 \times 3$, $5\times 5$, and $7 \times 7$, 
(ii)~lower and higher thresholds: $0.14$ and $0.22$, resp.
When compared to the edge points obtained using distances, ROA produces noisier results with higher false alarm results.
ROA improves when applied to the reduced resolution data set, in agreement with the fact that this method is especially designed to segment smooth targets.

In the following we discuss the effect of reducing the resolution on the distances using two actual images.

\begin{figure}[hbt]
	\centering
	\includegraphics[width=.85\linewidth]{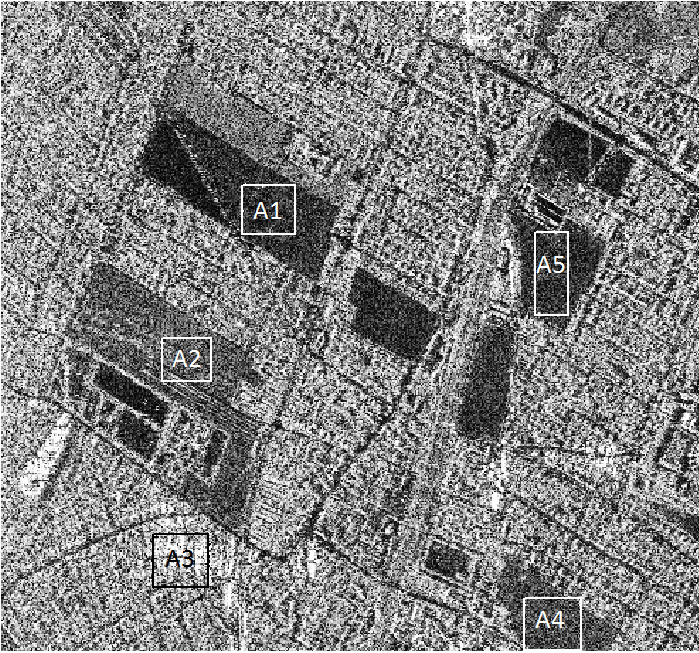}
	\caption{Regions from E-SAR data used to extract the proposed features. }
	\label{fig:realData}
\end{figure}

Fig.~\ref{fig:realData} presents the samples used to estimate the parameters, and to then measure GD and TD between them. 
We identified five samples:
A1 and A5 correspond to pasture regions, 
A2 and A4 areas are forest regions, 
and A3 corresponds to an urban region. 
We estimated the texture parameter $\alpha$ of the $\mathcal{G}_I^0$ distribution in both full and reduced resolution;
denote the estimates $\widehat{\alpha}$ and $\widehat{\alpha}_{\text{RR}}$ (reduced resolution), respectively.
These estimates are shown in Table~\ref{table_example}, and help in the identification of the types of land cover because they are related to the area roughness.
The table also shows the size of each sample in full resolution (the reduced resolution samples are one fourth of the original size).

These estimates are consistent with the land cover types visually identified.
The estimates reduce with the resolution in most cases; this is in agreement with the fact that the reduced resolution data were obtained by smoothing the observations.
This change reflects on changes in the distances.

\begin{table}[hbt]
	\centering
	\caption{Estimates of ${\alpha}$ in areas from the image in Fig.~\ref{fig:realData}.}
	\label{table_example}
	\begin{tabular}{@{}crrr@{}}
		\toprule
		Sample & Sample Size   & $\widehat{\alpha}$  & $\widehat{\alpha}_{\text{RR}}$ \\ \midrule
		A1  & $1360$ & $-6.09$  & $-11.53$  \\
		A2  & $1225$ & $-9.72$ & $-20.00$ \\
		A3  & $2088$ & $-1.01$ & $-1.00$ \\
		A4  & $1824$ & $-2.75$ & $-4.27$ \\
		A5  & $1152$  & $-11.51$ & $-20.00$ \\ \bottomrule
	\end{tabular}
\end{table}

Table~\ref{table_gedesicaMV} presents the distances between regions in both full and reduced resolution. 
	
\begin{table}[hbt]
	\centering
	\caption{Distances between models for actual data}
	\label{table_gedesicaMV}
	\begin{tabular}{@{}cccrrrr@{}}
		\toprule
		& Sample  & & A2    & A3    & A4    & A5      \\ \midrule
\multirow{8}{*}{\rotatebox{90}{Full Resolution}}
&\multirow{4}{*}{GD}
		& A1     & $0.467$  & $1.807$ & $0.794$ & $0.636$   \\
&		& A2     & $0 $    & $2.274$ & $1.261$ & $ 0.169$   \\
&		& A3     &       & $0 $    & $1.013$ & $ 2.443 $  \\
&		& A4     &       &       & $0$     & $1.430$  \\ \cmidrule{2-7}
&\multirow{4}{*}{TD}
		& A1     & $0.097$ & $0.840$ & $0.248$ & $ 0.170$   \\
&		& A2     & $0 $    & $ 1.097$ & $0.515$ & $0.014$   \\
&		& A3     &       & $0 $    & $0.367$ & $ 1.182 $  \\
&		& A4     &       &       & 0     & $0.617$   \\ \midrule
\multirow{8}{*}{\rotatebox{90}{Reduced Resolution}}
&\multirow{4}{*}{GD}
		& A1     & $2.277$ & $12.503$ & $4.004$ & $2.277$   \\
&		& A2     & $0 $    & $14.781$ & $6.282$ &   $0 $  \\
&		& A3     &       & $0 $    & $ 8.498$ & $ 14.781$   \\
&		& A4     &       &       & $0 $    & $6.282$  \\ \cmidrule{2-7}
&\multirow{4}{*}{TD}
		& A1     & $0.238$ & $ 1.524$ & $ 0.584$ & $ 0.238$   \\
&		& A2     & $0$     & $ 1.743$ & $1.061$ & $0 $  \\
&		& A3     &       & $0$     & $ 0.854 $ & $ 1.743$   \\
&		& A4     &       &       & $0 $    & $ 1.061$   \\ \bottomrule
	\end{tabular}
\end{table}

Table~\ref{table_gedesicaMV} shows that the largest and smallest distances correspond to areas with very different texture and to two smooth areas, respectively.
This indicates that region discrimination by means of both GD and TD is possible, as they behave as expected.
The distances change when the resolution changes, but the largest and smallest values of both GD and TD are still between A3 and A5, and A2 and A5, respectively, which indicates that the capability of texture discrimination is not affected by changes in the image resolution.

Table~\ref{table_gedesicaMV} also reveals that the GD is more sensitive to differences in the texture parameter values, allowing a finer discrimination between pasture and forest areas.
The TD correctly discriminates regions having very different texture, such as A3 (extreme textured area) and A5 (textureless area) but in the case of A1 and A2 which correspond to a textureless area and a moderate textured area, respectively, TD looses the ability to separate: its value is almost zero. 
This implicates an advantage of GD over TD. 

\begin{figure*}[hbt]
	\centering
	\subfigure[Distances between models for full resolution data]{\includegraphics[width=.4\linewidth]{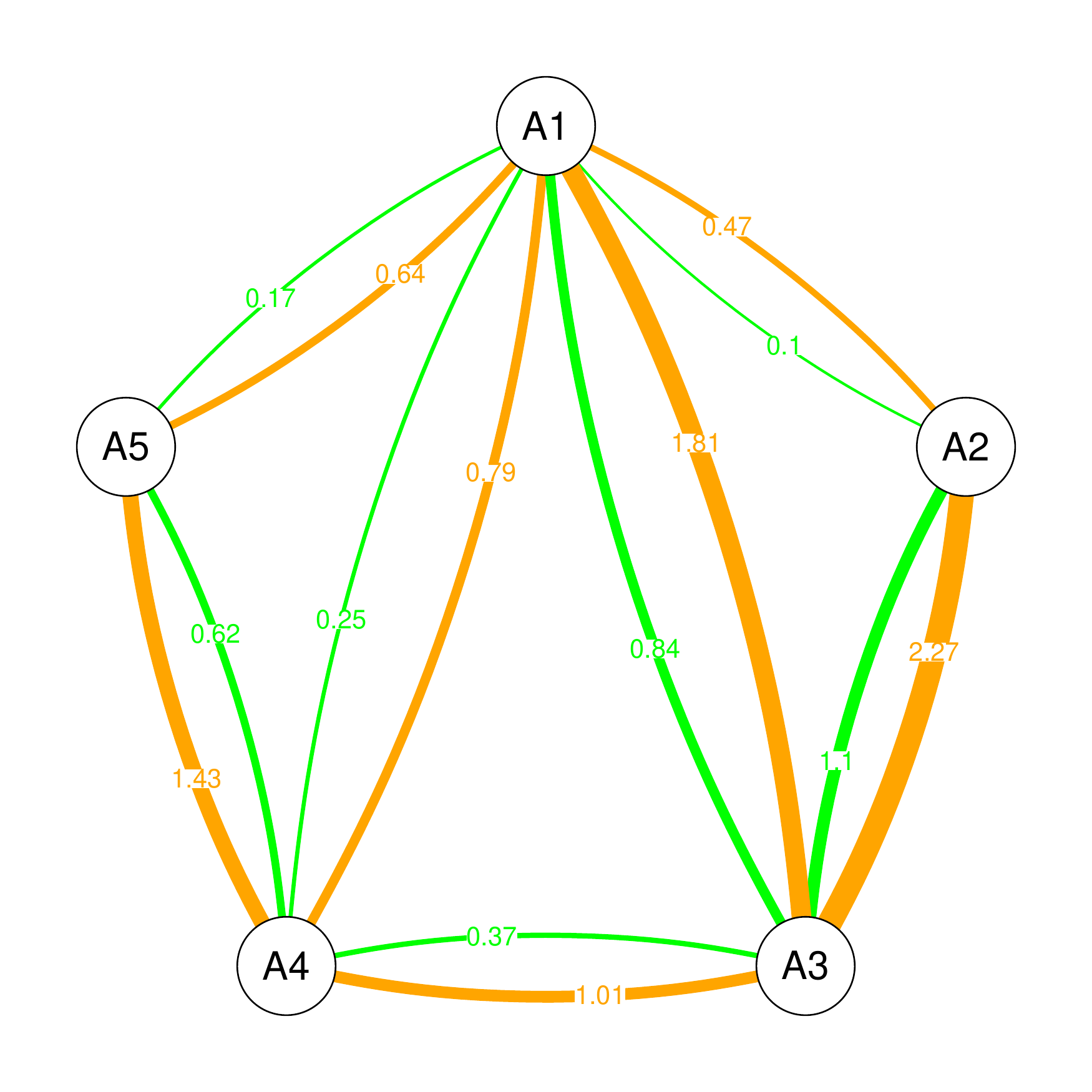}}
		\subfigure[Distances between models for reduced resolution data]{\includegraphics[width=.4\linewidth]{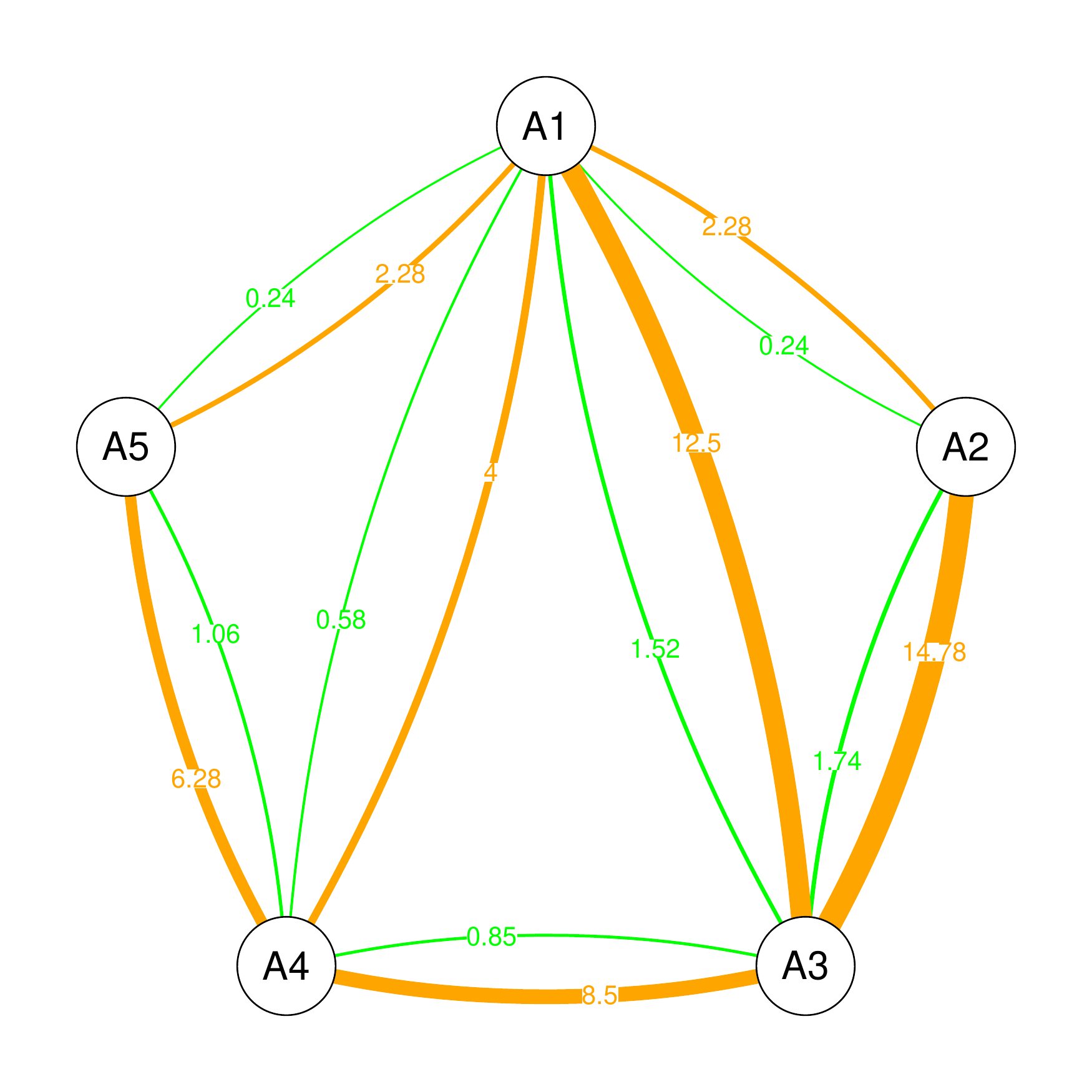}}
	\caption{Illustration distances between samples of true data in full and reduced resolution: GD in orange, and TD in green.}
	\label{fig:IlustracionTablasDistancias}
\end{figure*}

Figure~\ref{fig:IlustracionTablasDistancias} illustrates the distances between A1 and the other regions in both full and reduced resolution. 
GD and TD are represented with orange and green colors, respectively. 
The line thickness is proportional to the distances between region models. 

\begin{figure}[hbt]
	\centering
	\includegraphics[width=.6\linewidth]{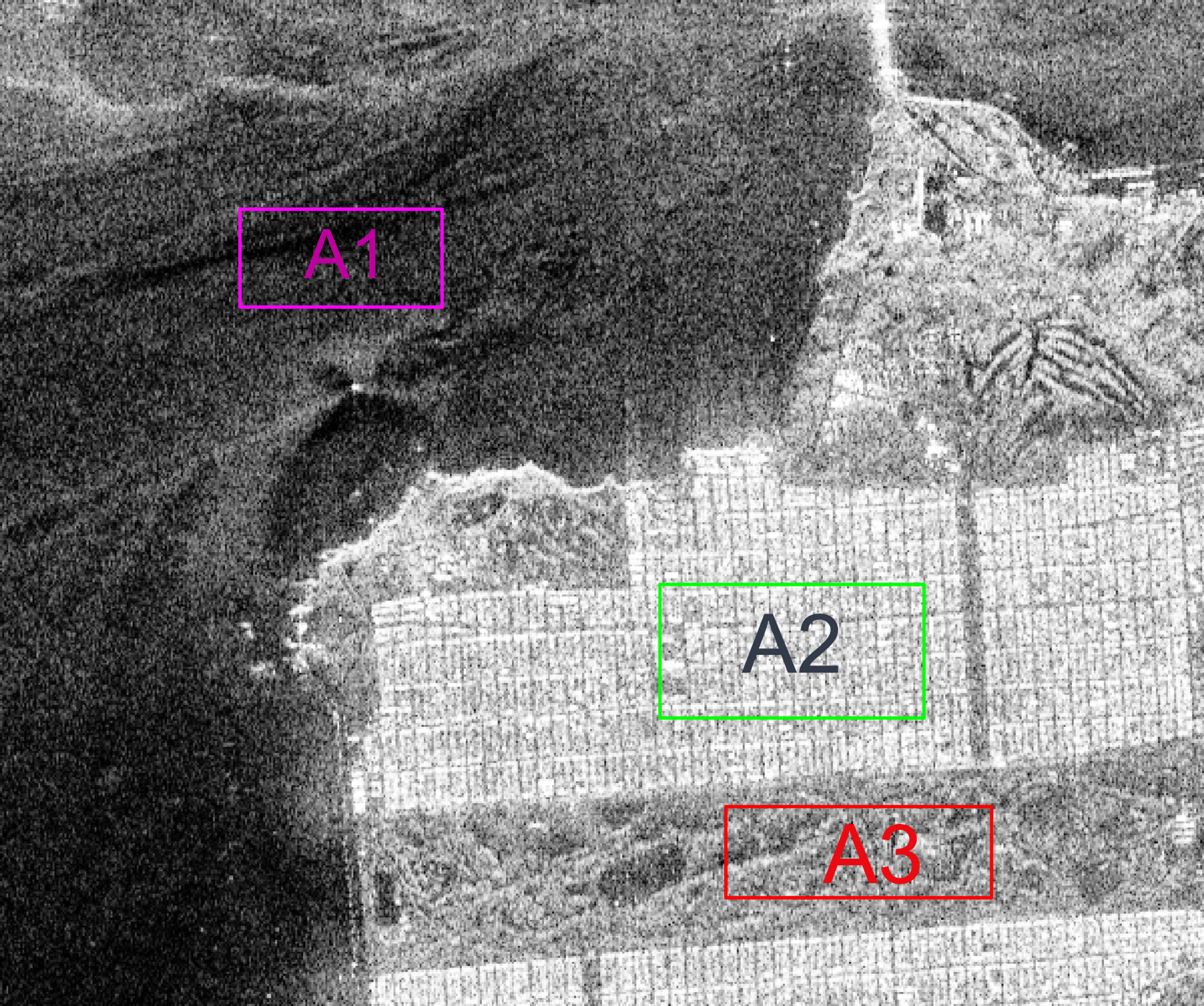}
	\caption{AIRSAR data over San Francisco, USA.}
		\label{fig:SanFra}
\end{figure}

Figure~\ref{fig:SanFra} shows an AIRSAR image over San Francisco, USA, HH polarization in intensity, with the equivalent number of looks $\widehat L=2$.  
We identified three samples: 
A1 corresponding to water, 
A2 an urban region, 
and A3 a forest area. 
The same procedure was performed as before for obtaining an image with reduced resolution.
Table~\ref{table_example_SanFra} shows the estimates of the $\alpha $ parameter for each region and the sample size, for both resolutions. 
It can be seen that the estimates agree with the visual interpretation of the classes.

\begin{table}[hbt]
	\centering
	\caption{Estimates of ${\alpha}$ in samples from Fig.~\ref{fig:SanFra}.}
	\label{table_example_SanFra}
	\begin{tabular}{@{}crrr@{}}
		\toprule
		Sample & Sample Size   & $\widehat{\alpha}$ & $\widehat{\alpha}_{\text{RR}}$ \\ \midrule
		A1  & $12282$ & $-11.307$ & $-11.488$ \\ 
		A2  & $8266$ & $-2.016$ & $-2.558$ \\
		A3  & $7878$ & $-5.746$ & $-6.485$\\ \bottomrule
	\end{tabular}
\end{table}

Table~\ref{table_gedesicaSanFra} presents the GD and the TD between regions. 
This example highlights the difficulty TD has to distinguish regions with a similar roughness. 
It can be seen that the TD between A1 and A3 and between A2 and A3 are almost the same in both resolutions. 

\begin{table}[hbt]
	\centering
	\caption{Distances between models for actual data}
	\label{table_gedesicaSanFra}
	\begin{tabular}{@{}cccrrrr@{}}
		\toprule
		& Sample  & & A2    & A3          \\ \midrule
\multirow{4}{*}{\rotatebox{90}{\small Full Res.}}
&\multirow{2}{*}{GD}
		& A1     & $6.86  $  & $2.74$   \\
&		& A2     & $0 $       & $4.12$  \\ \cmidrule{2-5}
&\multirow{2}{*}{TD}
		& A1     & $1.14 $ & $0.32$      \\
&		& A2     & $0 $      & $ 0.58$      \\ \midrule
\multirow{4}{*}{\rotatebox{90}{\small Red. Res.}}
&\multirow{2}{*}{GD}
		& A1     & $8.13$ & $3.22$    \\
&		& A2     & $0 $    & $4.91$    \\ \cmidrule{2-5}
&\multirow{2}{*}{TD}
		& A1     & $1.29$ & $ 0.38$   \\
&		& A2     & $0$     & $ 0.69$   \\ \bottomrule
	\end{tabular}
\end{table}
 
These experiments show that the methodology is not affected if the image has reduced resolution. 
Furthermore, as the integral of the TD is calculated numerically,  the computational cost is higher and it may even not converge, which is another advantage of the GD over the TD.

\section{Conclusions and Future Work}
\label{conclu}

This work is dedicated to deriving the Geodesic Distance (GD) between $\mathcal G_I^0$ models, a result that fills a gap in the literature of contrast measures for this distribution.
This distance can be used as a feature for, among other applications, edge detection in speckled data. 

We derived a closed formula for the geodesic distance for $L=\{1,2\}$, 
and we provide the general expression that can be used in other cases relying to numerical integration.
The unknown parameters were estimated using maximum likelihood to grant
interesting asymptotic properties for the GD. 
We compared the Geodesic and Triangular (TD) distances, with favorable results for the first over the second.

It is possible to observe that, as the lower difference between $\alpha_1$ and $\alpha_2$ is, the lower value of $s(\alpha_{1},\alpha_{2})$ is, which indicates that this distance is an appropriate feature to measure the contrast between regions.
 
The results of applying the GD for edge detection in simulated images are excellent.
Computing GD requires about $70$ times less computational time than TD.

When dealing with actual data from two sensors and two different scenes, 
large values of the GD and TD come from areas of the image with different roughness, but TD is consistently smaller than GD, making more difficult the identification of such differences.

The effectiveness of using GD is not affected by the resolution, but going beyond $L=2$ increases the computational cost for its calculation since numerical methods are required.

The value of $\gamma$ does not affect the results because it is a scale parameter of the $\mathcal{G}_I^0$ distribution.

One may transform GD and TD into test statistics with the same asymptotic distribution and, with this, they become comparable and tools for hypothesis testing. 
The finite-size behavior of the tests, cf.\ Fig.~\ref{fig:p-valor}, suggests the need of corrections in order to make them closer to the asymptotic result, but we observed that GD converges faster than TD to the asymptotic critical value.

As future work, we will experiment with contaminated data to measure the robustness of GD. 

\appendix


Simulations were performed using the \texttt{R} language and environment for statistical computing version~3.3.~\cite{Rmanual}, in a computer with processor Intel\textcopyright \ Core\texttrademark, i7-4790K CPU \SI{4}{\giga\hertz}, \SI{16}{\giga\byte} RAM, System Type \SI{64}{\bit} operating system. 

The numerical integration required to evaluate the geodesic distance for $L\neq 1,2$ and the triangular distance was performed with the \texttt{adaptIntegrate} function from the \texttt{cubature} package.
This is an adaptive multidimensional integration over hypercubes. 

The Google~Maps link for the image of Figure~\ref{fig:realData} is
\url{https://goo.gl/maps/DCm6fBKcRy32}.

Codes and data are available upon request from the corresponding author.


\begin{IEEEbiography}[{\includegraphics[width=1in]{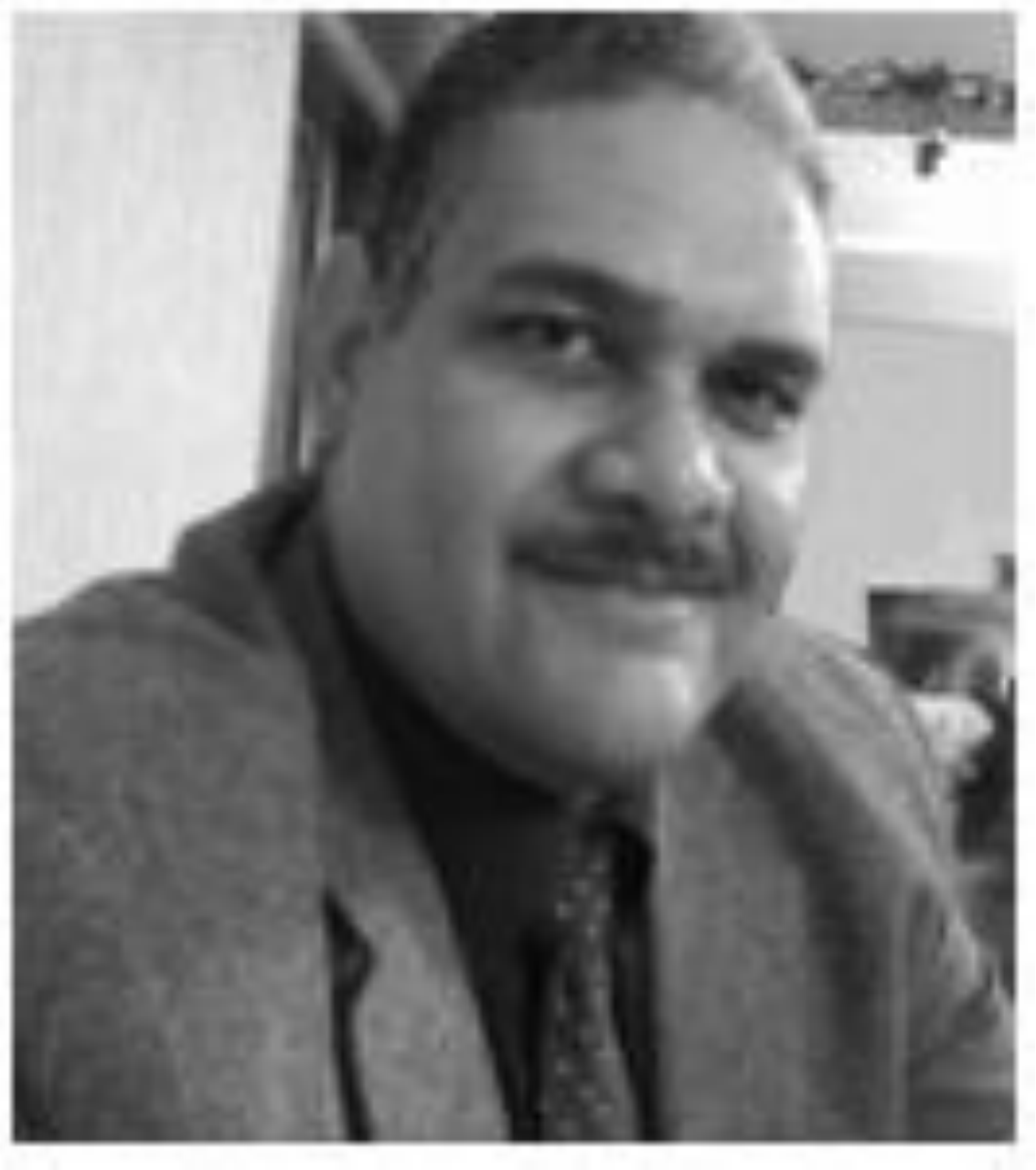}}]{Jos\'e Naranjo-Torres} 
is an aggregated professor at the Instituto Universitario de Tecnolog\'ia de Maracaibo (IUTM), Venezuela. He received the B.Sc. in Physics and M.Sc. in Geophysics degrees from the Universidad del Zulia, Venezuela. He is currently working toward the Ph.D. degree in the Universidad Nacional de General Sarmiento, Pcia. de Buenos Aires, Argentina. His research interests include statistical models for image processing and geophysics.
\end{IEEEbiography}

\begin{IEEEbiography}[{\includegraphics[width=1in]{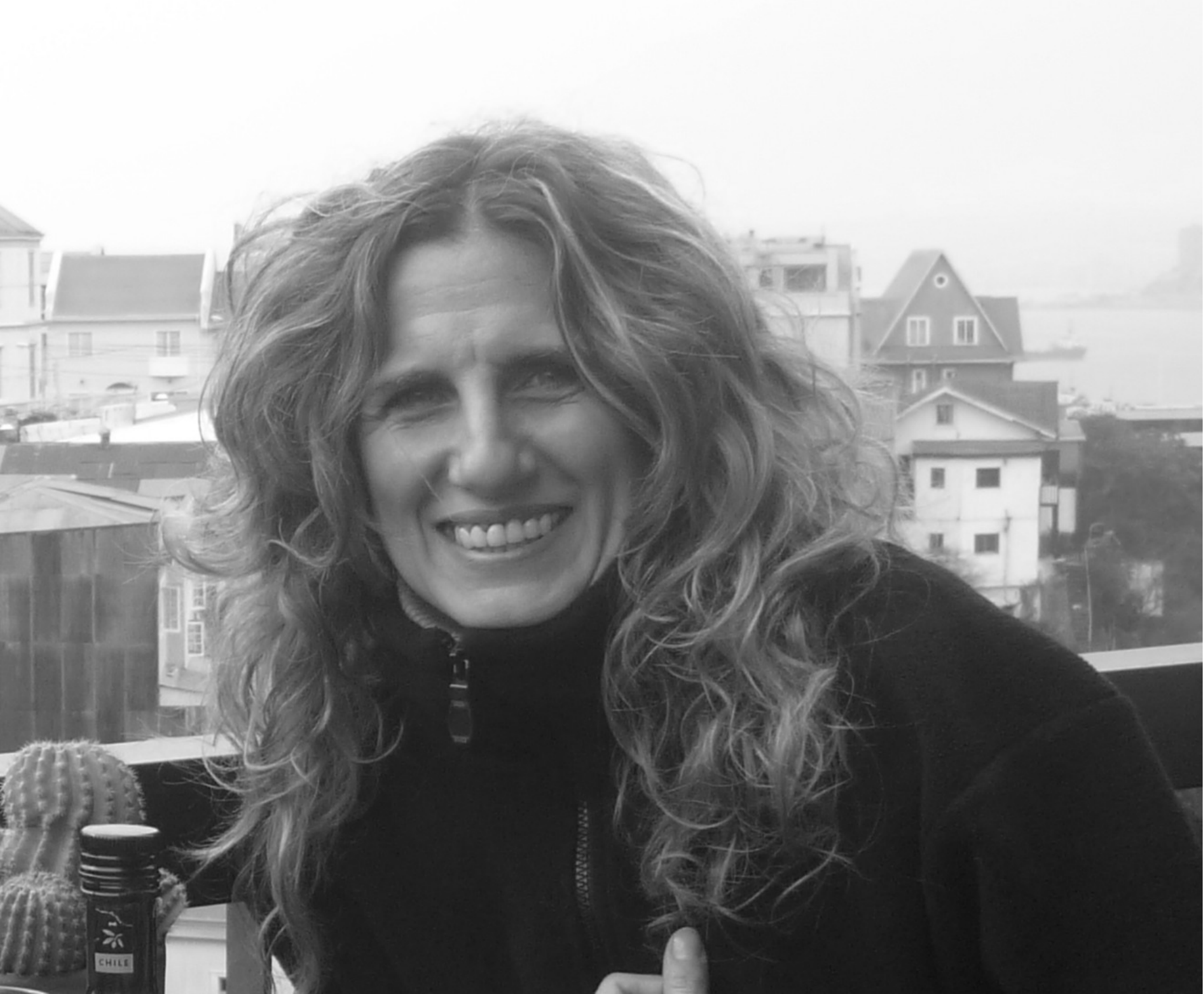}}]{Juliana Gambini} 
received the B.Sc. degree in Mathematics and the Ph.D. degree in Computer Science both from Universidad de Buenos Aires (UBA), Argentina.
She is currently Titular Professor at the Instituto Tecnol\'ogico de Buenos Aires (ITBA), Buenos Aires, Argentina, member of the Center for Computational Intelligence -- ITBA and a Titular Professor at Universidad Nacional de Tres de Febrero, Pcia. de Buenos Aires, Argentina.
Her research interests include SAR image processing, video processing and image recognition.
\end{IEEEbiography}

\begin{IEEEbiography}[{\includegraphics[width=1in]{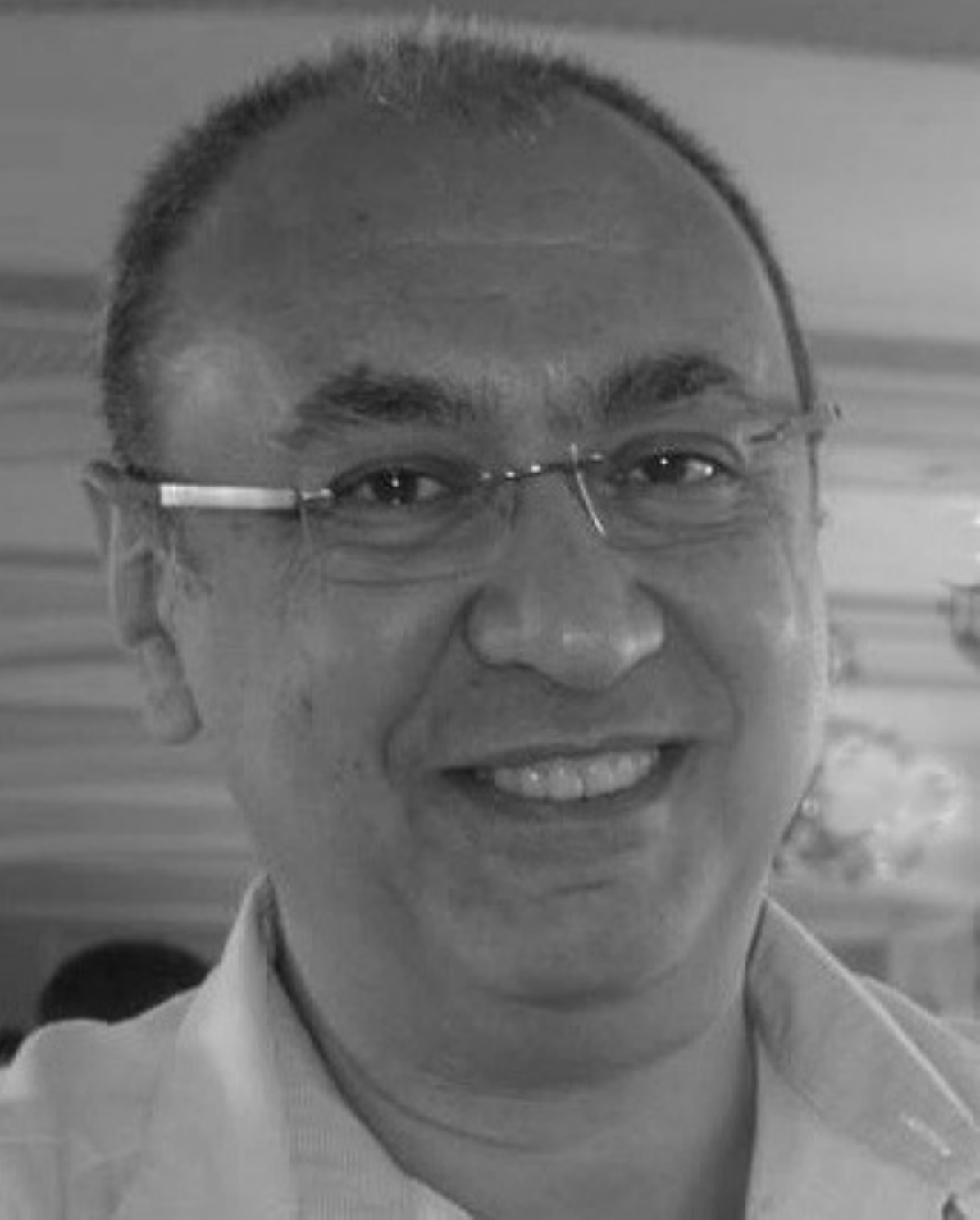}}]{Alejandro C.\ Frery} (S'92--SM'03)
received the B.Sc. degree in Electronic and Electrical Engineering from the Universidad de Mendoza, Mendoza, Argentina.
His M.Sc. degree was in Applied Mathematics (Statistics) from the Instituto de Matem\'atica Pura e Aplicada (IMPA, Rio de Janeiro) and his Ph.D. degree was in Applied Computing from the Instituto Nacional de Pesquisas Espaciais (INPE, S\~ao Jos\'e dos Campos, Brazil).
He is currently the leader of LaCCAN -- \textit{Laborat\'orio de Computa\c c\~ao Cient\'ifica e An\'alise Num\'erica}, Universidade Federal de Alagoas, Macei\'o, Brazil.
His research interests are statistical computing and stochastic modeling.
\end{IEEEbiography}
\vfill


\begin{thebibliography}{10}
\providecommand{\url}[1]{#1}
\csname url@samestyle\endcsname
\providecommand{\newblock}{\relax}
\providecommand{\bibinfo}[2]{#2}
\providecommand{\BIBentrySTDinterwordspacing}{\spaceskip=0pt\relax}
\providecommand{\BIBentryALTinterwordstretchfactor}{4}
\providecommand{\BIBentryALTinterwordspacing}{\spaceskip=\fontdimen2\font plus
\BIBentryALTinterwordstretchfactor\fontdimen3\font minus
  \fontdimen4\font\relax}
\providecommand{\BIBforeignlanguage}[2]{{%
\expandafter\ifx\csname l@#1\endcsname\relax
\typeout{** WARNING: IEEEtran.bst: No hyphenation pattern has been}%
\typeout{** loaded for the language `#1'. Using the pattern for}%
\typeout{** the default language instead.}%
\else
\language=\csname l@#1\endcsname
\fi
#2}}
\providecommand{\BIBdecl}{\relax}
\BIBdecl

\bibitem{6049331}
T.~Esch, M.~Schmidt, M.~Breunig, A.~Felbier, H.~Taubenb?ck, W.~Heldens,
  C.~Riegler, A.~Roth, and S.~Dech, ``Identification and characterization of
  urban structures using {VHR} {SAR} data,'' in \emph{IEEE International on
  Geoscience and Remote Sensing Symposium (IGARSS)}, 2011, pp. 1413--1416.

\bibitem{5764715}
Q.~Wu, R.~Chen, H.~Sun, and Y.~Cao, ``Urban building density detection using
  high resolution {SAR} imagery,'' in \emph{Joint Urban Remote Sensing Event},
  2011, pp. 45--48.

\bibitem{6352307}
C.~D. Storie, J.~Storie, and G.~S. de~Salmuni, ``Urban boundary extraction
  using 2-component polarimetric {SAR} decomposition,'' in \emph{IEEE
  International Geoscience and Remote Sensing Symposium (IGARSS)}, 2012, pp.
  5741--5744.

\bibitem{6235981}
F.~Dell'Acqua and P.~Gamba, ``Remote sensing and earthquake damage assessment:
  Experiences, limits, and perspectives,'' \emph{Proceedings of the IEEE}, vol.
  100, no.~10, pp. 2876--2890, 2012.

\bibitem{6242371}
M.~Sato, S.~W. Chen, and M.~Satake, ``Polarimetric {SAR} analysis of tsunami
  damage following the march 11, 2011 {E}ast {J}apan earthquake,''
  \emph{Proceedings of the IEEE}, vol. 100, no.~10, pp. 2861--2875, 2012.

\bibitem{6205771}
Y.~Yamaguchi, ``Disaster monitoring by fully {P}olarimetric {SAR} data acquired
  with {ALOS-PALSAR},'' \emph{Proceedings of the IEEE}, vol. 100, no.~10, pp.
  2851--2860, 2012.

\bibitem{6353568}
S.~W. Chen and M.~Sato, ``Tsunami damage investigation of built-up areas using
  multitemporal spaceborne full polarimetric {SAR} images,'' \emph{IEEE
  Transactions on Geoscience and Remote Sensing}, vol.~51, no.~4, pp.
  1985--1997, 2013.

\bibitem{7502166}
W.~Sun, L.~Shi, J.~Yang, and P.~Li, ``Building collapse assessment in urban
  areas using texture information from postevent {SAR} data,'' \emph{IEEE
  Journal of Selected Topics in Applied Earth Observations and Remote Sensing},
  vol.~9, no.~8, pp. 3792--3808, 2016.

\bibitem{6946812}
C.~Fu, Y.~Chen, L.~Tong, M.~Jia, L.~Tan, and X.~Ji, ``Road damage information
  extraction using high-resolution {SAR} imagery,'' in \emph{IEEE Geoscience
  and Remote Sensing Symposium (IGARSS)}, 2014, pp. 1836--1838.

\bibitem{7559238}
M.~Barber, C.~L\'opez-Mart\'inez, and F.~Grings, ``Assessment of {L}-{B}and
  {SAR} polarimetry for soil and crop monitoring,'' in \emph{European
  Conference on Synthetic Aperture Radar (EUSAR)}, June 2016, pp. 1--4.

\bibitem{6352427}
R.~T. Melrose, R.~T. Kingsford, and A.~K. Milne, ``Using radar to detect
  flooding in arid wetlands and rivers,'' in \emph{IEEE International
  Geoscience and Remote Sensing Symposium (IGARSS)}, 2012, pp. 5242--5245.

\bibitem{6350840}
D.~Velotto, S.~Lehner, A.~Soloviev, and C.~Maingot, ``Analysis of oceanic
  features from dual-polarization high resolution {X}-band {SAR} imagery for
  oil spill detection purposes,'' in \emph{IEEE International Geoscience and
  Remote Sensing Symposium (IGARSS)}, 2012, pp. 2841--2844.

\bibitem{Cropdiscriminationbasedonpolarimetric}
S.~Chen, Y.~Li, and X.~Wang, ``Crop discrimination based on polarimetric
  correlation coefficients optimization for {P}ol{SAR} data,''
  \emph{International Journal of Remote Sensing}, vol.~36, no.~16, pp.
  4233--4249, 2015.

\bibitem{SurfaceRoughnessScatteringHighResolutionSAR}
S.-E. Park, L.~Ferro-Famil, S.~Allain, and E.~Pottier, ``Surface roughness and
  microwave surface scattering of high-resolution imaging radar,'' \emph{IEEE
  Geoscience and Remote Sensing Letters}, vol.~12, no.~4, pp. 756--760, Apr
  2015.

\bibitem{martindenicolas2014}
M.~Presa, J.~Amores, D.~Mata~Moya, and J.~B·rcena~Humanes, ``Statistical
  analysis of {SAR} sea clutter for classification purposes,'' \emph{Remote
  Sensing}, vol.~6, no.~10, pp. 9379--9411, 2014.

\bibitem{Farrouki2005}
A.~Farrouki and M.~Barkat, ``Automatic censoring {CFAR} detector based on
  ordered data variability for nonhomogeneous environments,'' \emph{IEE
  Proceedings Radar, Sonar and Navigation}, vol. 152, no.~1, pp. 43--51, 2005.

\bibitem{35954}
J.-S. Lee and I.~Jurkevich, ``Segmentation of {SAR} images,'' \emph{IEEE
  Transactions on Geoscience and Remote Sensing}, vol.~27, no.~6, pp. 674--680,
  1989.

\bibitem{1198658}
R.~Fj{\o}rtoft, Y.~Delignon, W.~Pieczynski, M.~Sigelle, and F.~Tupin,
  ``Unsupervised classification of radar images using hidden {M}arkov chains
  and hidden {M}arkov random fields,'' \emph{IEEE Transactions on Geoscience
  and Remote Sensing}, vol.~41, no.~3, pp. 675--686, 2003.

\bibitem{7394295}
P.~X. Vu, N.~T. Duc, and V.~V. Yem, ``Application of statistical models for
  change detection in {SAR} imagery,'' in \emph{International Conference on
  Communications, Management and Telecommunications}, 2015, pp. 239--244.

\bibitem{Gambiniijrs}
J.~Gambini, M.~Mejail, J.~Jacobo-Berlles, and A.~Frery, ``Feature extraction in
  speckled imagery using dynamic {B}-spline deformable contours under the
  $\mathcal{G}^0$ model,'' \emph{International Journal of Remote Sensing},
  vol.~27, no.~22, pp. 5037--5059, 2006.

\bibitem{Frery97}
A.~C. Frery, H.-J. M{\"u}ller, C.~C.~F. Yanasse, and S.~J.~S. Sant'Anna, ``A
  model for extremely heterogeneous clutter,'' \emph{IEEE Transactions on
  Geoscience and Remote Sensing}, vol.~35, no.~3, pp. 648--659, 1997.

\bibitem{6985522}
Y.~Wang, T.~L. Ainsworth, and J.~S. Lee, ``On characterizing high-resolution
  {SAR} imagery using kernel-based mixture speckle models,'' \emph{IEEE
  Geoscience and Remote Sensing Letters}, vol.~12, no.~5, pp. 968--972, 2015.

\bibitem{MejailJacoboFreryBustos:IJRS}
M.~Mejail, J.~C. Jacobo-Berlles, A.~C. Frery, and O.~H. Bustos,
  ``Classification of {SAR} images using a general and tractable multiplicative
  model,'' \emph{International Journal of Remote Sensing}, vol.~24, no.~18, pp.
  3565--3582, 2003.

\bibitem{Gao2010}
G.~Gao, ``Statistical modeling of {SAR} images: A survey,'' \emph{Sensors},
  vol.~10, no.~1, pp. 775--795, 2010.

\bibitem{GambiniSC08}
J.~Gambini, M.~Mejail, J.~Jacobo-Berlles, and A.~C. Frery, ``Accuracy of edge
  detection methods with local information in speckled imagery,''
  \emph{Statistics and Computing}, vol.~18, no.~1, pp. 15--26, 2008.

\bibitem{ClassificationPolSARSegmentsMinimizationWishartDistances}
W.~B. Silva, C.~C. Freitas, S.~J.~S. Sant'Anna, and A.~C. Frery,
  ``Classification of segments in {PolSAR} imagery by minimum stochastic
  distances between {W}ishart distributions,'' \emph{IEEE Journal of Selected
  Topics in Applied Earth Observations and Remote Sensing}, vol.~6, no.~3, pp.
  1263--1273, 2013.

\bibitem{ClassificationFullyPolSARDiffusion-Reaction}
L.~Gomez, L.~Alvarez, L.~Mazorra, and A.~C. Frery, ``Classification of complex
  {W}ishart matrices with a diffusion-reaction system guided by stochastic
  distances,'' \emph{Philosophical Transactions of the Royal Society A:
  Mathematical, Physical and Engineering Sciences}, vol. 373, no. 2056, 2015.

\bibitem{6377288}
A.~C. Frery, R.~J. Cintra, and A.~D.~C. Nascimento, ``Entropy-based statistical
  analysis of {P}ol{SAR} data,'' \emph{IEEE Transactions on Geoscience and
  Remote Sensing}, vol.~51, no.~6, pp. 3733--3743, 2013.

\bibitem{5208318}
A.~D.~C. Nascimento, R.~J. Cintra, and A.~C. Frery, ``Hypothesis testing in
  speckled data with stochastic distances,'' \emph{IEEE Transactions on
  Geoscience and Remote Sensing}, vol.~48, no.~1, pp. 373--385, 2010.

\bibitem{EdgeDetectionDistancesEntropiesJSTARS}
A.~D.~C. Nascimento, M.~M. Horta, A.~C. Frery, and R.~J. Cintra, ``Comparing
  edge detection methods based on stochastic entropies and distances for
  {PolSAR} imagery,'' \emph{IEEE Journal of Selected Topics in Applied Earth
  Observations and Remote Sensing}, vol.~7, no.~2, pp. 648--663, 2014.

\bibitem{gambini2015}
J.~Gambini, J.~Cassetti, M.~Lucini, and A.~Frery, ``Parameter estimation in
  {SAR} imagery using stochastic distances and asymmetric kernels,'' \emph{IEEE
  Journal of Selected Topics in Applied Earth Observations and Remote Sensing},
  vol.~8, no.~1, pp. 365--375, 2015.

\bibitem{NonparametricEdgeDetectionSpeckledImagery}
E.~Gir\'on, A.~C. Frery, and F.~Cribari-Neto, ``Nonparametric edge detection in
  speckled imagery,'' \emph{Mathematics and Computers in Simulation}, vol.~82,
  pp. 2182--2198, 2012.

\bibitem{7562415}
R.~H. Nobre, F.~A.~A. Rodrigues, R.~C.~P. Marques, J.~S. Nobre, J.~F. S.~R.
  Neto, and F.~N.~S. Medeiros, ``{SAR} image segmentation with {R}eny's
  entropy,'' \emph{IEEE Signal Processing Letters}, vol.~23, no.~11, pp.
  1551--1555, 2016.

\bibitem{6947510}
X.~Huang, P.~Huang, L.~Dong, H.~Song, and W.~Yang, ``Saliency detection based
  on distance between patches in polarimetric {SAR} images,'' in \emph{IEEE
  International Geoscience and Remote Sensing Symposium (IGARSS)}, 2014, pp.
  4572--4575.

\bibitem{TorresPolarimetricFilterPatternRecognition}
L.~Torres, S.~J.~S. Sant'Anna, C.~C. Freitas, and A.~C. Frery, ``Speckle
  reduction in polarimetric {SAR} imagery with stochastic distances and
  nonlocal means,'' \emph{Pattern Recognition}, vol.~47, pp. 141--157, 2014.

\bibitem{6410023}
E.~Erten, ``Glacier velocity estimation by means of a polarimetric similarity
  measure,'' \emph{IEEE Transactions on Geoscience and Remote Sensing},
  vol.~51, no.~6, pp. 3319--3327, 2013.

\bibitem{7326080}
W.~Yang, H.~Song, G.~S. Xia, and C.~L\'opez-Mart\'inez, ``Dissimilarity
  measurements for processing and analyzing {P}ol{SAR} data: A survey,'' in
  \emph{IEEE International Geoscience and Remote Sensing Symposium (IGARSS)},
  2015, pp. 1562--1565.

\bibitem{Conradsen2003}
K.~Conradsen, A.~A. Nielsen, J.~Schou, and H.~Skriver, ``A test statistic in
  the complex {W}ishart distribution and its application to change detection in
  polarimetric {SAR} data,'' \emph{IEEE Transactions on Geoscience and Remote
  Sensing}, vol.~41, no.~1, pp. 4--19, 2003.

\bibitem{Schou2003}
J.~Schou, H.~Skriver, A.~A. Nielsen, and K.~Conradsen, ``{CFAR} edge detector
  for polarimetric {SAR} images,'' \emph{IEEE Transactions on Geoscience and
  Remote Sensing}, vol.~41, no.~1, pp. 20--32, 2003.

\bibitem{ChenWang2012}
S.~W. Chen, X.~S. Wang, and M.~Sato, ``Pol{I}n{SAR} complex coherence
  estimation based on covariance matrix similarity test,'' \emph{IEEE
  Transactions on Geoscience and Remote Sensing}, vol.~50, no.~11, pp.
  4699--4710, 2012.

\bibitem{7010344}
S.~Tu, Y.~Li, and Y.~Su, ``Ratio- and distribution-metric-based active contours
  for {SAR} image segmentation,'' in \emph{Fifth International Conference on
  Intelligent Control and Information Processing (ICICIP)}, 2014, pp. 227--232.

\bibitem{7517404}
L.~Mascolo, J.~M. Lopez-Sanchez, F.~Vicente-Guijalba, F.~Nunziata,
  M.~Migliaccio, and G.~Mazzarella, ``A complete procedure for crop phenology
  estimation with {P}ol{SAR} data based on the complex wishart classifier,''
  \emph{IEEE Transactions on Geoscience and Remote Sensing}, vol.~54, no.~11,
  pp. 6505--6515, 2016.

\bibitem{raey1945}
C.~R. Rao, ``Information and the accuracy attainable in the estimation of
  statistical parameters,'' \emph{Bulletin of Calcutta Mathematical Society},
  vol.~37, pp. 81--91, 1945.

\bibitem{raey1992}
------, ``\BIBforeignlanguage{English}{Information and the accuracy attainable
  in the estimation of statistical parameters},'' in
  \emph{\BIBforeignlanguage{English}{Breakthroughs in Statistics}}, ser.
  Springer Series in Statistics, S.~Kotz and N.~L. Johnson, Eds.\hskip 1em plus
  0.5em minus 0.4em\relax New York: Springer, 1992, pp. 235--247.

\bibitem{ISI:000302346300002}
G.~Verdoolaege and P.~Scheunders, ``On the geometry of multivariate
  {G}eneralized {G}aussian models,'' \emph{Journal of Mathematical Imaging and
  Vision}, vol.~43, no.~3, pp. 180--193, 2012.

\bibitem{ISI:A1995RC74900013}
M.~L. Menendez, D.~Morales, L.~Pardo, and M.~Salicru, ``Statistical test based
  on the geodesic distances,'' \emph{Applied Mathematics Letters}, vol.~8,
  no.~{1}, pp. 65--69, 1995.

\bibitem{AtkinsonMitchell1981}
C.~Atkinson and A.~F.~S. Mitchell, ``Rao's distance measure,''
  \emph{Sankhy{\=a}: The Indian Journal of Statistics, Series A (1961-2002)},
  vol.~43, pp. 345--365, 1981.

\bibitem{Berkane1997}
M.~Berkane, K.~Oden, and P.~Bentler, ``Geodesic estimation in elliptical
  distributions,'' \emph{Journal of Multivariate Analysis}, vol.~63, pp.
  35--46, 1997.

\bibitem{Reverter2003155}
F.~Reverter and J.~Oller, ``Computing the {R}ao distance for {G}amma
  distributions,'' \emph{Journal of Computational and Applied Mathematics},
  vol. 157, no.~1, pp. 155--167, 2003.

\bibitem{bombrun2011}
L.~Bombrun, Y.~Berthoumieu, N.-E. Lasmar, and G.~Verdoolaege, ``Multivariate
  texture retrieval using the geodesic distance between elliptically
  distributed random variables,'' in \emph{18th IEEE International Conference
  on Image Processing (ICIP)}, 2011, pp. 3637--3640.

\bibitem{IOPORT.10013799}
A.~D. El~Maliani, M.~El~Hassouni, Y.~Berthoumieu, and D.~Aboutajdine, ``Color
  texture classification method based on a statistical multi-model and geodesic
  distance,'' \emph{Journal of Visual Communication and Image Representation},
  vol.~25, no.~7, pp. 1717--1725, 2014.

\bibitem{5946541}
L.~Bombrun, N.-E. Lasmar, Y.~Berthoumieu, and G.~Verdoolaege, ``Multivariate
  texture retrieval using the {SIRV} representation and the geodesic
  distance,'' in \emph{IEEE International Conference on Acoustics, Speech and
  Signal Processing (ICASSP)}, 2011, pp. 865--868.

\bibitem{FreitasFreryCorreia:Environmetrics:03}
C.~C. Freitas, A.~C. Frery, and A.~H. Correia, ``The polarimetric {G}
  distribution for {SAR} data analysis,'' \emph{Environmetrics}, vol.~16,
  no.~1, pp. 13--31, 2005.

\bibitem{6049750}
H.~Zhong, Q.~Xie, L.~Jiao, and S.~Wang, ``Water/land segmentation for {SAR}
  images based on geodesic distance,'' in \emph{IEEE International Geoscience
  and Remote Sensing Symposium (IGARSS)}, 2011, pp. 2661--2664.

\bibitem{Naranjo2015}
J.~Naranjo-Torres, J.~Gambini, and A.~C. Frery, ``Region discrimination in
  {SAR} imagery using the geodesic distance between {GI0} distributions,'' in
  \emph{IEEE 5th Asia-Pacific Conference on Synthetic Aperture Radar (APSAR)},
  2015, pp. 573--577.

\bibitem{OntheApplicationsofDivergenceTypeMeasuresinTestingStatisticalHypothesessalicru}
M.~Salicr\'u, D.~Morales, M.~L. Men\'endez, and L.~Pardo, ``On the applications
  of divergence type measures in testing statistical hypotheses,''
  \emph{Journal of Multivariate Analysis}, vol.~51, no.~2, pp. 372--391, 1994.

\bibitem{QuartulliDatcu:04}
M.~Quartulli and M.~Datcu, ``Stochastic geometrical modelling for built-up area
  understanding from a single {{SAR}} intensity image with meter resolution,''
  \emph{IEEE Transactions on Geoscience and Remote Sensing}, vol.~42, no.~9,
  pp. 1996--2003, 2004.

\bibitem{Broyden65}
C.~G. Broyden, ``A class of methods for solving nonlinear simultaneous
  equations,'' \emph{Mathematics of Computation}, vol.~19, pp. 577--593, 1965.

\bibitem{6600931}
O.~Besson and Y.~I. Abramovich, ``On the {F}isher information matrix for
  multivariate elliptically contoured distributions,'' \emph{IEEE Signal
  Processing Letters}, vol.~20, no.~11, pp. 1130--1133, 2013.

\bibitem{7551770}
G.~Marti, S.~Andler, F.~Nielsen, and P.~Donnat, ``Optimal transport vs.
  {F}isher-{R}ao distance between copulas for clustering multivariate time
  series,'' in \emph{IEEE Statistical Signal Processing Workshop (SSP)}, 2016,
  pp. 1--5.

\bibitem{Horn1998}
R.~Horn, ``{E-SAR}: The experimental airborne {L}/{C}-band {SAR} system of
  {DFVLR},'' in \emph{IEEE International Geoscience and Remote Sensing
  Symposium (IGARSS)}, vol.~2.\hskip 1em plus 0.5em minus 0.4em\relax IEEE
  Press, 1988, pp. 1025--1026.

\bibitem{touzi88}
R.~Touzi, A.~Lopes, and P.~Bousquet, ``A statistical and geometrical edge
  detector for {SAR} images,'' \emph{IEEE Trans. Geosci. Remote Sensing},
  vol.~26, no.~11, pp. 764--773, 1988.

\bibitem{Rmanual}
\BIBentryALTinterwordspacing
{R Core Team}, \emph{R: A Language and Environment for Statistical Computing},
  R Foundation for Statistical Computing, Vienna, Austria, 2016. [Online].
  Available: \url{https://www.R-project.org/}
\BIBentrySTDinterwordspacing

\end{thebibliography}
\end{document}